\documentclass[11pt,x11names]{article}
\usepackage{setspace}
\usepackage{misc/acl}

\usepackage{times}
\usepackage[table,x11names]{xcolor}
\usepackage{graphicx}
\usepackage[T1]{fontenc}
\usepackage[utf8]{inputenc}
\usepackage{microtype}
\usepackage{booktabs}
\usepackage{todonotes}
\usepackage{amsmath,amssymb}
\usepackage{cleveref}
\usepackage{xspace}
\usepackage{soul} %
\usepackage{caption}
\usepackage{multirow,multicol}
\usepackage{ulem}
\usepackage{float}
\usepackage{array}
\usepackage{bm}
\usepackage{xfrac}
\usepackage{tcolorbox} %
\usepackage[shortlabels]{enumitem}
\usepackage[outline]{contour}%
\usepackage{tikz}
\usepackage{hyperref}
\usepackage{pgfplots}
\pgfplotsset{compat=1.18}

\setlist[itemize]{noitemsep,left=0mm}

\usepackage{tabularx}

\crefname{lstlisting}{listing}{listings}
\Crefname{lstlisting}{Listing}{Listings}
\crefname{equ}{equation}{equations}
\Crefname{equ}{Equation}{Equations}
\Crefname{algorithm}{Algorithm}{Algorithms}
\crefname{example}{example}{examples}
\Crefname{example}{Example}{Examples}
\crefname{prompt}{prompt}{prompts}
\Crefname{prompt}{Prompt}{Prompts}
\DeclareCaptionType{example}[Example][List of examples]
\DeclareCaptionType{prompt}[Prompt][List of prompts]
\DeclareCaptionType{equ}[Equation][List of equations]

\usepackage{marginnote}

\definecolor{TodoColor}{rgb}{1,0.7,0.6}
\definecolor{TodoColor2}{rgb}{0.7,0.7,0.9}
\definecolor{TodoColor3}{rgb}{0.5,0.8,0.5}

\newcommand{\ACref}[1]{Appendix \Cref{#1}}

\makeatletter\def\Hy@Warning#1{}\makeatother
\let\svthefootnote\thefootnote
\newcommand\blankfootnote[1]{%
  \let\thefootnote\relax\footnotetext{#1}%
  \let\thefootnote\svthefootnote%
}

\graphicspath{{img/}}

\title{
    How Important is `Perfect' English for Machine Translation Prompts?
}

\usepackage{authblk}
\usepackage[misc]{ifsym}
\author{
\bf Patrícia Schmidtová$^{*, 1}$ 
\qquad
Niyati Bafna$^{*, 2}$ 
\qquad
Seth Aycock$^{*, 3}$
\\
\bf
Gianluca Vico$^1$
\quad
Wiktor Kamzela$^4$
\quad
Katharina Hämmerl$^{\dagger, 5,6}$
\quad 
Vilém Zouhar$^{\dagger, 7}$
\\\\
$^1${Charles University, Faculty of Mathematics and Physics}\,\,
$^2${Johns Hopkins University}\\
$^3${University of Amsterdam}\,\,
$^4${Poznań University of Technology}\,\,
$^5${TU Munich}\\
$^6${Munich Center for Machine Learning}\,\,
$^7${ETH Zurich}
}

\begin{document}

\maketitle

\maketitle

\begin{abstract}
Large language models (LLMs) achieve state-of-the-art performance in machine translation, but are also known to be sensitive to errors in user prompts.
Given these models are overwhelmingly trained on and respond best to prompts in standard English, this may affect the quality of LLM outputs for second language English speakers as well as real-world lay users, with potentially disproportionate effects on the former.
We explore this effect by modeling and synthetically producing a range of error types exhibited by such users, motivated by studies of L2 English, and quantifying their impact on LLM performance. 
We work with two related tasks: machine translation and machine translation evaluation.
We find that LLMs-as-MTs are brittle to natural spelling-inspired errors but not to errors on the phrasal level.
However, the variance in quality caused by these errors is lower than the variance over the initial prompt choice, suggesting that perfect English for a given prompt is less important than choosing a good prompt. 
Since lay users and L2 speakers may naturally use non-optimal prompts as well as display imperfect language skills, our work calls for increasing the resilience of model performance to both these phenomena to best serve a diverse user base, both from a robustness and fairness perspective.

\end{abstract}

\blankfootnote{
 \hspace{-1mm}$^*$Equal contribution,
 $^\dagger$Co-last authors.\\
 \null\hspace{6mm}Corresponding author \href{mailto:schmidtova@ufal.mff.cuni.cz}{schmidtova@ufal.mff.cuni.cz}
 }

\footnotetext[0]{ 
We release the \href{https://huggingface.co/datasets/niyatibafna/imperfect_english_prompts}{translations dataset} in 3 language pairs from 6 state-of-the-art (closed \& open) models and 7 error augmenters to enable further research into the effects of prompt quality on LLM performance; as well as the \href{https://github.com/patuchen/imperfect_english_prompts}{code}.
}

\section{Introduction}
\label{sec:introduction}
Large language models (LLMs) have recently dominated machine translation benchmarks \citep{kocmi-etal-2024-findings}.
These models are known to work best with English prompts, even for tasks in other languages \citep{dey2024betteraskenglishevaluation}, and are notoriously sensitive to errors in their prompts \citep[][inter alia]{qiang-etal-2024-prompt}.

\begin{figure}[t]
\centering
\includegraphics[width=1\linewidth]{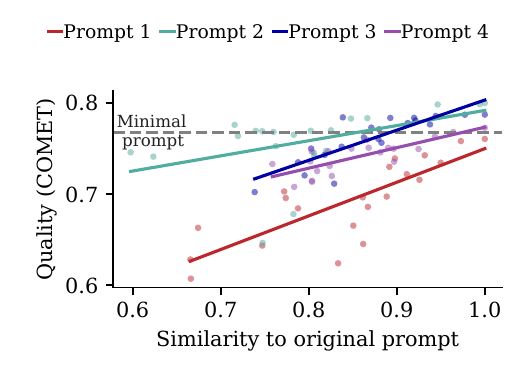}

\vspace{-3mm}
\caption{Changing model performance, as measured by COMET score (y-axis), across all error-augmented (orthographic errors) prompts and all models.
The similarity of each error-augmented prompt to the original is measured by the inner product of their sentence embeddings (x-axis).}
\label{fig:quality-v-sim-all}
\end{figure}

Research publications and model evaluation setups tend to use well-crafted and tuned prompts in correctly spelled, grammatical, and `standard' language.
However, users, i.e., the actual target audience of the models, constitute a diverse user base, including language learners and L2 speakers, as well as lay people in real-world conditions.
Not only are users unlikely to tune their prompts, they may also exhibit a variety of errors in their prompts, stemming both from lack of proficiency in English as well as from usage in real-world conditions and natural style variation. 
Thus, the evaluation mode is misaligned with the way the models are used and evaluations in research might misrepresent the true model performance in the wild.

Our work aims to fill this practice-evaluation gap.
While past works explore general prompt noise robustness, in this work we model the effect of specific patterns of errors from users on LLM performance.
We evaluate LLM robustness to user errors in prompts on two machine translation-related tasks: machine translation itself, and LLM-based machine translation evaluation \citep{Kocmi23-LargeLanguageModels}.

Given various error classes of interest, we simulate real-world LLM usage by synthesising these types of errors in the user prompts in a controlled manner with varying intensities.
This allows us to quantify and compare the impact of different error types on task performance as well as on the ability of the model to generate outputs in the intended language (on-targetness). 
We also perform a small qualitative analysis of the resulting performance degradation and error modes.
We focus specifically on errors in the \textit{user prompt}, rather than other parts of the input, such as the system prompt.
This mimics the typical user/researcher use-case, who would not usually have access to the system prompt of state-of-the-art LLMs.

\paragraph{Findings.}
Through a large-scale quantitative evaluation of the effect of seven error profiles across three language pairs, six state-of-the-art models, followed by qualitative analysis, we find that:
\begin{itemize}[left=0mm,noitemsep]
\item \textbf{Error type matters}: Spelling errors have the greatest impact on LLM performance, while sentence-level simplifications and other phrase-level phenomena typical of L2 speakers or lazy users do not significantly degrade performance.
\item \textbf{Prompt choice dominates}: The initial prompt choice has greater influence on performance than the vast majority of realistic user errors.
\item \textbf{Errors reduce instruction-following but not translation quality}: Errors in prompts primarily affect models' ability to follow instructions (e.g. to avoid redundant text alongside translations or produce off-target translations) rather than their core translation capabilities, with LLMs demonstrating surprising and unpredictable robustness to severe errors.

\end{itemize}

\noindent
Similar findings also hold true for the sibling task of translation quality estimation: Lower-quality prompts show a weakly detrimental effect on the automatic quality assessment, as meta-evaluated by system-level correlation with human judgments.

\section{Related Work}

\paragraph{LLMs for Machine Translation.}
General-purpose decoder-only LLMs have demonstrated state-of-the-art performance in machine translation with zero- and few-shot prompts \citep{kocmi-etal-2024-findings}.
However, LLMs may refuse to answer or generate redundant text surrounding the translation which adversely affects automatic evaluation \citep{briakou2024implicationsverbosellmoutputs}.
Further, performance has been shown to vary considerably depending on the chosen prompt \citep{bawden-yvon-2023-investigating}.

While LLMs show strong translation performance with zero-shot prompting \citep{Hendy23-HowGoodAre}, this is particularly true for explicitly multilingual models such as EuroLLM \citep{Martins24-EuroLLMMultilingualLanguage}.
Both fine-tuning \citep{Xu23-ParadigmShiftMachine} and instruction-tuning on the translation task can further boost performance \citep{alves-etal-2023-steering1}.
For example, TowerLLM \citep{rei-etal-2024-tower}, which is instruction-tuned for multilingual translation and related tasks, achieved leading results on the WMT24 general translation task \citep{kocmi-etal-2024-findings}.

\paragraph{Robustness of LLMs.}

Robustness of language models has been explored in the context of adaptation to low-resource settings and user-generated text.
\citet{srivastava-chiang-2025-calling} model multiple types of variation in input segments automatically and focus on variations in English, and \citet{bafna-etal-2024-evaluating} focus on dialectal variation.

\citet{belinkov2018synthetic} diagnosed NMT models to be sensitive to both synthetic and natural errors in the input text.
More recently, \citet{peters-martins-2025-translation} found GPT-3.5 to be surprisingly resilient to synthetic errors.
They also looked at the input segment and not the user prompt, applying synthetic typos in 10-100\% of input tokens. 

Relatively little work has addressed errors in the prompt specifically.
\citet{zhu2024promptrobust} generate `adversarial' prompts containing possible typos and semantic errors, as generated by several different tools.
Additionally, while they cover a number of tasks, these are mostly classification tasks.
\citet{gonen2024demystifyingpromptslanguagemodels} showed that prompt effectiveness is correlated with its perplexity under an LLM, implying that deviant or non-standard prompts are likely to do worse.
This motivates our work, which quantifies the effect of natural deviation and non-standardness as exhibited by real-word users.

In contrast to previous work, this paper examines the impact of \emph{naturalistic} error types with \emph{varying error intensities} on the prompt (rather than input segments).

\paragraph{LLM-as-a-judge for Translation.}
LLMs have been shown to be effective evaluators of models' instruction-following abilities \citep{Zheng23-JudgingLLMasaJudgeMTBench}, and have since been successfully applied to translation evaluation.
\citet{kocmi-federmann-2023-gemba, Kocmi23-LargeLanguageModels} introduce GEMBA, a prompt-based metric using GPT-4 to produce direct assessments (DA), multidimensional quality metric (MQM) analysis, or error span annotations (ESA; \citealp{kocmi-etal-2024-error}).
In this work, we use the zero-shot GEMBA-DA prompt which achieves high system-level correlations with human judgments for both reference-based evaluation and reference-free quality estimation, competitive with fine-tuned metrics \citep{freitag-etal-2023-results, freitag-etal-2024-llms} such as CometKiwi \citep{Rei22-CometKiwiISTUnbabel2022} and XCOMET-QE \citep{guerreiro-etal-2024-xcomet}.
Improvements to LLM-based quality estimation are observed with chain-of-thought prompting for error analysis \citep{Lu24-ErrorAnalysisPrompting} and fine-tuning on human judgments, which boosts poor segment-level correlations of LLMs-as-judges \citep{Fernandes23-DevilErrorsLeveraging}.
\citet{Huang24-LostSourceLanguage} investigate the effect of including source and references on evaluation performance.
Closest to this part of our work, \citet{qian-etal-2024-large} examine the effect of prompt formatting for LLM judges for translation.

\begin{figure}[t]
    \centering
    \includegraphics[width=1\linewidth]{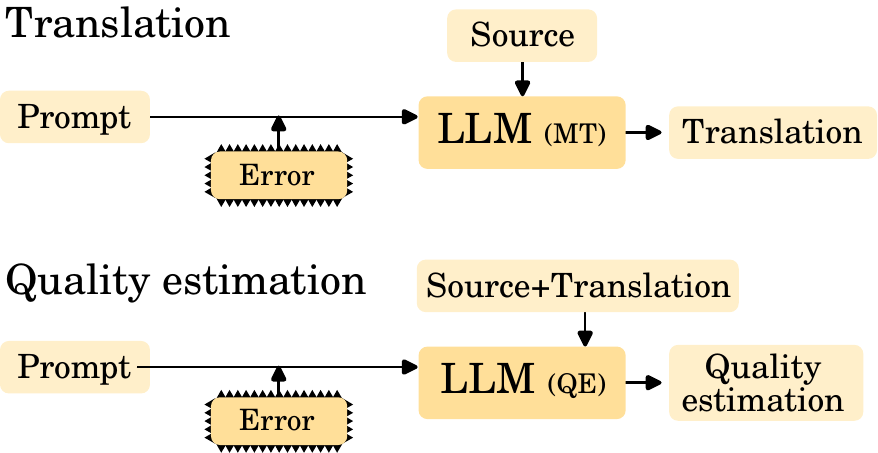}
    \caption{\textbf{Top:} Machine translation pipeline. The original prompt is augmented with an error, then filled with source language, target language, and source sentence, before being translated by an LLM. \textbf{Bottom:} Quality estimation pipeline. Translations are evaluated using the GPT Estimation Metric Based Assessment (GEMBA).}
    \label{fig:highlevel_pipeline}
\end{figure}

\section{Methods}
\label{sec:noise-func}

Our aim is to model the effect of a range of \textit{error types} made by users on LLM performance in a controlled setting.
To this end, we design \textit{error augmenters} to imitate each of several different error types of interest, motivated by real-world use. Each error augmenter allows us to model varying intensities of the respective error type. 
Users may also of course exhibit several error types simultaneously: thus, we also study the effect of logical compositions of particular error types that mimic users of particular profiles of interest. 

A full description of the individual error augmenters is given in \Cref{app:noise-functions}.
See examples of the error-augmented prompts in \Cref{tab:noised_prompts}.

\subsection{Modeling Error Profiles}

Our error types model the following scenarios:

\begin{itemize}[left=0mm,noitemsep]
    
    \item \textbf{Natural orthographic errors} (spelling errors) due to imperfect proficiency is modeled by our orthographic error augmenter (\ref{app:subsec:ortho-noise}).
    This introduces character-level perturbations with a probability $p$, modeling documented errors from L1 and L2 speakers \citep{cook1997l2} (e.g. confusion between particular sets of letters and common vowel sequence transpositions) as well as random typos.
    We manually choose a range for $p \in [0,0.4]$ as representing a natural spectrum for the intensity of this type of error, and generate error-augmented prompts for $10$ uniformly spaced values of $p$ within this range.
    We may imagine that the latter parts of the range represent less proficient L2 writers, including L1 and L2 children.
    \item \textbf{Phonetic errors} (\ref{app:subsec:phonetic-llm}) supplements the above error augmenter by introducing spelling errors motivated by phonetic guesses. 
    \item \textbf{Phrasal errors} (\ref{app:subsec:phrasal-noise}) mimics phrasal substitutions and simplifications as made by beginner and intermediate speakers of English. We instructed an LLM to generate candidates for two intensity levels, and manually post-edited and modified the resulting prompts for naturalness.
    \item \textbf{Register errors} (\ref{app:subsec:register-noise}) also operates on a phrase level, and deals in the register, or the level of formality or casualness, that the user applies. We generated over two intensity levels, similarly as with phrasal errors.
    \item \textbf{Low-proficiency writers, such as L2 learners} of English, presumably commit lexical/phrasal errors as well as spelling errors.
    We model this as a combined scenario by applying orthographic errors (with the same settings as above) over both levels of the phrasal error augmenter.
    This results in $10 \cdot 2 = 20$ error-augmented versions per prompt.
    Different compositions of the two error augmenters can be imagined to represent the diversity of proficiency in English, i.e. users with syntactic proficiency but imperfect spelling or vice versa.
    \item \textbf{Lazy users} use informal registers, and presumably also make spelling errors. As above, we compose the orthographic error augmenter in the selected range over both levels of the register error augmenter to generate prompts with varied error intensities.
    \item \textbf{Uniform errors.} We also apply a control character-level error augmenter that samples perturbations uniformly at random, rather than by user-inspired patterns such as the above (\ref{app:subsec:random-noise}). This helps us contrast model tolerance to the resulting ``unnatural'' perturbations against more natural error patterns as simulated above. The error augmenter creates random character substitutions with a probability $p$; we use $10$ choices of $p$ uniformly spaced between $[0,1]$. 

\end{itemize}

\noindent
For all scenarios involving the uniform and orthographic error augmenters, we generate $20$ erroneous prompts per parametrization.

\subsection{Error Intensity and Sampling}
Recall that we want to obtain error-augmented prompts over a range of error intensities in order to measure the effect of errors on LLM performance. 
We would also like to compare the effect of different error \textit{types} or profiles.
However, the intensity of various error types scales differently.
For example,
different levels of spelling errors do not correspond directly to different levels of phrasal errors.
To enable a consistent interpretation of impact of various error types on performance, as well as permit cross-error comparisons, we measure LLM performance against a `unit error' introduced by an error type, where `unit error' is measured by the resultant distance (or similarity) of the error-augmented prompt to the original prompt. 
We use the following two measures to capture this similarity:

\begin{itemize}
    \item \textbf{chrF} between the base prompt and the error-augmented version gives a surface measure of prompt similarity, where a lower chrF score is associated with a higher error intensity.
    \item \textbf{Inner product of embeddings} of the base prompt and the error-augmented prompts gives a more semantic measure of deviation of the error-augmented prompts from the original. The embeddings are derived from \texttt{all-MiniLM-L6-v2} from SentenceTransformers \citep{reimers-2019-sentence-bert}.
\end{itemize}

\noindent
By using various parameterizations per scenario as described above, we obtain error-augmented prompts that model our error profiles of interest over a range of error intensities as per the above measures.
For every error profile, we can now observe the correlation between LLM task performance given an error-augmented prompt and the amount of deviation in that prompt from the original. 
Intuitively, this allows us to answer the question of which error profiles cause the most damage, given an equivalent amount of perturbation to the prompt. 

In practice, we measure this correlation over discrete buckets of increasing error intensity. Given a bucket, the prompt used for a particular input is sampled randomly over error-augmented prompts in that bucket.
This provides stable estimates of model performance by reducing vulnerability to outlier prompts.

\section{Experimental Settings}

\paragraph{Prompts.}
We choose four zero-shot prompts used by LLM-based systems at the WMT24 General Translation task \citep{kocmi-etal-2024-findings} as our base prompts. 
As a sanity check, we include results for an additional minimalistic baseline that was shown to perform well by \citet{zhang_prompting}. 
We do not apply any errors to this baseline.
See \Cref{tab:base-prompts} for the full baseline prompts, \Cref{tab:noised_prompts} for examples of perturbed prompts, and \Cref{sec:implementation_details} for implementation details.

\paragraph{Setup.}
We select two closed-source API models and four open-weight models:
\begin{itemize}[left=4mm]
    \item GPT-4o-mini \citep{OpenAI24-GPT4oMiniAdvancing}
    \item Gemini-2.0-flash \citep{Google24-IntroducingGemini20}
    \item Llama-3.1-8B-Instruct \citep{Dubey24-LlamaHerdModels}
    \item Qwen2.5-7B-Instruct \citep{Qwen25-Qwen25TechnicalReport}
    \item EuroLLM-9B-Instruct \citep{Martins24-EuroLLMMultilingualLanguage}
    \item TowerInstruct-7B-v0.2 \citep{rei-etal-2024-tower}
\end{itemize}

\noindent
The models are selected so that they support the languages used for the experiments, and, for the open-weight models, that we can run them on our infrastructure. 
See \ACref{tab:model-selection} for the list of models considered and their supported languages.

We use language pairs present in WMT 2024 \citep{kocmi-etal-2024-findings}, specifically: Czech-Ukrainian, German-English, and English-Chinese.
Qwen officially supports only English, while the other models either officially support %
or empirically show good performance on these languages (i.e., by taking part in WMT24).
For each language pair, we randomly choose 500 segments, which is close to the total number in the test set.
For evaluation, we use ChrF \citep{popovic-2015-chrf} and \href{https://huggingface.co/Unbabel/wmt22-comet-da}{COMET$_\mathrm{22}^\mathrm{DA}$} \citep{rei-etal-2022-comet}.
We rely on both because COMET is known to struggle on out-of-distribution translations \citep{zouhar-etal-2024-pitfalls}.
See \Cref{app:eval-details} for the evaluation settings.

\paragraph{Quality Estimation with GEMBA.}

We use GPT-4o-mini for consistency with our translation experiments. 
We use two base prompts: GEMBA-DA (quality estimation Prompt 1, \citealp{Kocmi23-LargeLanguageModels}) and TMU-HIT's WMT24 quality estimation prompt (QE Prompt 2, \citealp{sato-etal-2024-tmu}); full prompts are shown in \ACref{tab:qe-base-prompts}.
We test on Czech-Ukrainian, German-English, and English-Chinese, as for translation.
We meta-evaluate the quality estimation performance by computing system and segment level Pearson correlations with human scores on submitted WMT24 systems.
We follow a strict setup with no retries; when GEMBA fails to output a correctly formatted score, we set the score for that segment to 0.
We limit experiments on the quality estimation task to orthographic errors on the two base prompts, both to maintain a realistic scenario and to limit costs.

\begin{table}[t]
    \small  \centering
    \begin{tabular}{@{\hspace{1mm}}p{7.3cm}@{\hspace{1mm}}}
    \toprule
        \textbf{Prompt 3:}
        Translate this from \{src\_lang\} to \{tgt\_lang\}:\textbackslash n \{src\_lang\}: \{src\_text\}\textbackslash n \{tgt\_lang\}: \\[2em]
        \textbf{Orthographic (0.1)}
        Trranslate ti from \{src\_lang\} too \{tgt\_lang:\}: \textbackslash n \{src\_lang\}: \{src\_text\} \textbackslash n \{tgt\_lang\}:
        \\
        \textbf{Orthographic (0.4):}
        Tranzlate dhiss from \{src\_lang\} to \{tgt\_lang\}: \textbackslash n \{src\_lang\}: \{src\_text\} \textbackslash n \{tgt\_lang\}:
        \\

        \textbf{Lexical/Phrasal (1):}
        Make this text in \{tgt\_lang\} from \{src\_lang\}: \textbackslash n \{src\_lang\}: \{src\_text\} \textbackslash n \{tgt\_lang\}:
        \\
        \textbf{Lexical/Phrasal (2):}
        You translate this text to \{tgt\_lang\} fromm \{src\_lang\}:  \textbackslash n \{src\_lang\}: \{src\_text\}  \textbackslash n \{tgt\_lang\}:
        \\
        \textbf{Phonetic}:
        Tranzlate thees from \{src\_lang\} to \{tgt\_lang\}:  \textbackslash n \{src\_lang\}: \{src\_text\}  \textbackslash n \{tgt\_lang\}:
        \\
        \textbf{Register (1)}:
        \{tgt\_lang\} version of this pls:  \textbackslash n \{src\_lang\}: \{src\_text\}  \textbackslash n \{tgt\_lang\}:
        \\
        \textbf{Register (2)}:
        change lang \{src\_lang\} -> \{tgt\_lang\}:  \textbackslash n \{src\_lang\}: \{src\_text\}  \textbackslash n \{tgt\_lang\}: \\
    \bottomrule      
    \end{tabular}
    \caption{Various types and levels (denoted in parentheses) of errors applied to Prompt 3.}
    \label{tab:noised_prompts}
\end{table}

\section{Results and Discussion}

\subsection{Prompt Choice is Critical}
First, we look at the effect of the prompt choice itself.
\Cref{fig:quality-v-sim-all} shows the changes in translation quality depending on the semantic similarity to the original prompt (averaged across all models).
Clearly, all four base prompts are affected by applying error.
There are also large differences in performance between the base prompts on their own, showing that prompt choice matters for state-of-the-art performance.

The `minimal' prompt yields a reasonable performance but stays behind the best base prompts.
Its key benefit compared to the other prompts is the fact that the minimal prompt is essentially impossible to make mistakes with:
Using an otherwise well-performing prompt with many errors leads to much worse performance, and may be worse than using a generally poorly-performing prompt, or the minimal prompt.
At the same time, a generally poorly-performing prompt without errors can still perform worse than a generally better-performing prompt with a few errors.
These observations reinforce the intuition that both prompt choice and correctness matter for best results.

Our prompts also respond differently to various error types. 
\Cref{tab:boron} shows Pearson correlations of translation quality with the prompt similarity to the base prompt, per error type and per prompt.
We discuss the various error types further in \Cref{subsec:compare-noise}.
Prompts 3 and 4 appear more resilient to noising overall.
For instance, for prompt 3, there is no correlation between error intensity and translation quality when using realistic orthographic errors.
This is likely due to their short length.
Because we preserve the critical variables of source and target language, as well as the input segment, noising the rest of these short prompts introduces less confusion than when noising the more complex prompts 1 and 2.
Additionally, prompt 4 even seems to benefit from phrasal errors, perhaps because its base form underperforms.

\newcommand{\microplot}[1]{
\begin{tikzpicture}
  \begin{axis}[
    xmin=0.6, xmax=1.0,
    ymin=0.5, ymax=0.9,
    ticks=none,
    axis x line=bottom,
    axis y line=left,
    enlargelimits=false,
    height=2.5cm,
    width=2.5cm,
  ]
    \addplot[only marks, mark=*, mark size=0.8pt, color=black] coordinates {
      #1
    };
  \end{axis}
\end{tikzpicture}
}
\begin{table*}[t]
\small \centering
\begin{tabular}{lllll@{\hspace{7mm}}l}\toprule
\bf Error augmenter & \bf Prompt 1 & \bf Prompt 2 & \bf Prompt 3 & \bf Prompt 4 & \bf All prompts\\
\midrule
\raisebox{2mm}{Uniform error} & \cellcolor{Brown3!43}\hspace{-2mm}\raisebox{2mm}{\phantom{-} 0.86 }\microplot{
(0.98, 0.76)
(0.88, 0.73)
(0.91, 0.73)
(0.79, 0.74)
(0.85, 0.70)
(0.86, 0.62)
(0.76, 0.61)
(0.71, 0.52)
(0.68, 0.40)
(0.70, 0.49)
} & \cellcolor{Brown3!46}\hspace{-2mm}\raisebox{2mm}{\phantom{-} 0.92 }\microplot{
(0.94, 0.80)
(0.85, 0.78)
(0.83, 0.78)
(0.84, 0.76)
(0.76, 0.69)
(0.68, 0.64)
(0.70, 0.65)
(0.66, 0.51)
(0.65, 0.50)
} & \cellcolor{Brown3!34}\hspace{-2mm}\raisebox{2mm}{\phantom{-} 0.68 }\microplot{
(0.99, 0.79)
(0.87, 0.76)
(0.90, 0.77)
(0.93, 0.78)
(0.81, 0.77)
(0.76, 0.69)
(0.78, 0.73)
(0.85, 0.78)
(0.78, 0.64)
(0.79, 0.59)
} & \cellcolor{Brown3!31}\hspace{-2mm}\raisebox{2mm}{\phantom{-} 0.61 }\microplot{
(0.94, 0.75)
(0.85, 0.74)
(0.94, 0.76)
(0.75, 0.76)
(0.72, 0.68)
(0.92, 0.74)
(0.68, 0.71)
(0.72, 0.69)
(0.78, 0.63)
(0.70, 0.64)
} & \cellcolor{Brown3!38}\hspace{-2mm}\raisebox{2mm}{\phantom{-} 0.77 }\microplot{
(0.94, 0.75)
(0.85, 0.74)
(0.94, 0.76)
(0.75, 0.76)
(0.72, 0.68)
(0.92, 0.74)
(0.68, 0.71)
(0.72, 0.69)
(0.78, 0.63)
(0.70, 0.64)
(0.99, 0.79)
(0.87, 0.76)
(0.90, 0.77)
(0.93, 0.78)
(0.81, 0.77)
(0.76, 0.69)
(0.78, 0.73)
(0.85, 0.78)
(0.78, 0.64)
(0.79, 0.59)
(0.98, 0.76)
(0.88, 0.73)
(0.91, 0.73)
(0.79, 0.74)
(0.85, 0.70)
(0.86, 0.62)
(0.76, 0.61)
(0.71, 0.52)
(0.68, 0.40)
(0.70, 0.49)
(0.94, 0.80)
(0.85, 0.78)
(0.83, 0.78)
(0.84, 0.76)
(0.76, 0.69)
(0.68, 0.64)
(0.70, 0.65)
(0.66, 0.51)
(0.65, 0.50)
}\\
\raisebox{2mm}{Phonetic} & \cellcolor{Brown3!42}\hspace{-2mm}\raisebox{2mm}{\phantom{-} 0.84 }\microplot{
(0.67, 0.63)
(0.67, 0.66)
(0.75, 0.64)
(0.67, 0.61)
(0.77, 0.70)
(0.79, 0.68)
(0.86, 0.70)
(0.77, 0.70)
(0.87, 0.69)
(0.90, 0.74)
} & \cellcolor{Brown3!36}\hspace{-2mm}\raisebox{2mm}{\phantom{-} 0.72 }\microplot{
(0.80, 0.77)
(0.76, 0.77)
(0.78, 0.77)
(0.62, 0.74)
(0.74, 0.77)
(0.72, 0.76)
(0.72, 0.78)
(0.60, 0.75)
(0.75, 0.77)
(0.76, 0.75)
} & \cellcolor{Brown3!43}\hspace{-2mm}\raisebox{2mm}{\phantom{-} 0.86 }\microplot{
(0.94, 0.78)
(0.79, 0.73)
(0.80, 0.72)
(0.74, 0.70)
(0.84, 0.78)
(0.88, 0.77)
(0.80, 0.75)
(0.92, 0.78)
(0.86, 0.76)
(0.82, 0.74)
} & \cellcolor{Brown3!28}\hspace{-2mm}\raisebox{2mm}{\phantom{-} 0.57 }\microplot{
(0.90, 0.74)
(0.81, 0.73)
(0.76, 0.73)
(0.80, 0.75)
(0.85, 0.75)
(0.88, 0.75)
(0.87, 0.75)
(0.80, 0.74)
(0.78, 0.71)
(0.87, 0.76)
} & \cellcolor{Brown3!37}\hspace{-2mm}\raisebox{2mm}{\phantom{-} 0.75 }\microplot{
(0.90, 0.74)
(0.81, 0.73)
(0.76, 0.73)
(0.80, 0.75)
(0.85, 0.75)
(0.88, 0.75)
(0.87, 0.75)
(0.80, 0.74)
(0.78, 0.71)
(0.87, 0.76)
(0.94, 0.78)
(0.79, 0.73)
(0.80, 0.72)
(0.74, 0.70)
(0.84, 0.78)
(0.88, 0.77)
(0.80, 0.75)
(0.92, 0.78)
(0.86, 0.76)
(0.82, 0.74)
(0.67, 0.63)
(0.67, 0.66)
(0.75, 0.64)
(0.67, 0.61)
(0.77, 0.70)
(0.79, 0.68)
(0.86, 0.70)
(0.77, 0.70)
(0.87, 0.69)
(0.90, 0.74)
(0.80, 0.77)
(0.76, 0.77)
(0.78, 0.77)
(0.62, 0.74)
(0.74, 0.77)
(0.72, 0.76)
(0.72, 0.78)
(0.60, 0.75)
(0.75, 0.77)
(0.76, 0.75)
}\\
\raisebox{2mm}{Lazy user} & \cellcolor{Brown3!47}\hspace{-2mm}\raisebox{2mm}{\phantom{-} 0.94 }\microplot{
(1.00, 0.76)
(0.95, 0.77)
(0.92, 0.74)
(0.88, 0.72)
(0.83, 0.67)
(0.87, 0.68)
(0.78, 0.59)
(0.81, 0.65)
(0.76, 0.53)
(0.76, 0.53)
} & \cellcolor{Brown3!38}\hspace{-2mm}\raisebox{2mm}{\phantom{-} 0.77 }\microplot{
(0.99, 0.80)
(0.96, 0.80)
(0.71, 0.71)
(0.78, 0.77)
(0.80, 0.77)
(0.72, 0.75)
(0.71, 0.77)
(0.79, 0.72)
(0.57, 0.47)
(0.74, 0.69)
} & \cellcolor{Brown3!18}\hspace{-2mm}\raisebox{2mm}{\phantom{-} 0.36 }\microplot{
(0.76, 0.75)
(0.95, 0.79)
(0.84, 0.75)
(0.95, 0.79)
(0.90, 0.79)
(0.84, 0.74)
(0.84, 0.78)
(0.83, 0.71)
(0.83, 0.74)
(0.84, 0.60)
} & \cellcolor{Brown3!26}\hspace{-2mm}\raisebox{2mm}{\phantom{-} 0.52 }\microplot{
(0.82, 0.76)
(0.94, 0.76)
(0.78, 0.75)
(0.94, 0.77)
(0.88, 0.75)
(0.83, 0.74)
(0.81, 0.66)
(0.77, 0.64)
(0.84, 0.61)
(0.83, 0.65)
} & \cellcolor{Brown3!32}\hspace{-2mm}\raisebox{2mm}{\phantom{-} 0.65 }\microplot{
(0.82, 0.76)
(0.94, 0.76)
(0.78, 0.75)
(0.94, 0.77)
(0.88, 0.75)
(0.83, 0.74)
(0.81, 0.66)
(0.77, 0.64)
(0.84, 0.61)
(0.83, 0.65)
(0.76, 0.75)
(0.95, 0.79)
(0.84, 0.75)
(0.95, 0.79)
(0.90, 0.79)
(0.84, 0.74)
(0.84, 0.78)
(0.83, 0.71)
(0.83, 0.74)
(0.84, 0.60)
(1.00, 0.76)
(0.95, 0.77)
(0.92, 0.74)
(0.88, 0.72)
(0.83, 0.67)
(0.87, 0.68)
(0.78, 0.59)
(0.81, 0.65)
(0.76, 0.53)
(0.76, 0.53)
(0.99, 0.80)
(0.96, 0.80)
(0.71, 0.71)
(0.78, 0.77)
(0.80, 0.77)
(0.72, 0.75)
(0.71, 0.77)
(0.79, 0.72)
(0.57, 0.47)
(0.74, 0.69)
}\\
\raisebox{2mm}{Orthographic} & \cellcolor{Brown3!35}\hspace{-2mm}\raisebox{2mm}{\phantom{-} 0.71 }\microplot{
(1.00, 0.75)
(0.96, 0.74)
(0.99, 0.76)
(0.93, 0.75)
(0.88, 0.72)
(0.93, 0.74)
(0.89, 0.70)
(0.94, 0.72)
(0.84, 0.73)
} & \cellcolor{Brown3!29}\hspace{-2mm}\raisebox{2mm}{\phantom{-} 0.58 }\microplot{
(0.99, 0.80)
(0.96, 0.80)
(0.89, 0.80)
(0.90, 0.80)
(0.94, 0.80)
(0.87, 0.78)
(0.87, 0.75)
(0.86, 0.77)
(0.82, 0.77)
(0.84, 0.80)
} & \cellcolor{Blue3!1}\hspace{-2mm}\raisebox{2mm}{ -0.01 }\microplot{
(0.96, 0.79)
(0.96, 0.79)
(0.94, 0.79)
(0.86, 0.78)
(1.00, 0.79)
(0.79, 0.79)
(0.92, 0.78)
(0.92, 0.79)
(0.95, 0.79)
(0.85, 0.79)
} & \cellcolor{Brown3!34}\hspace{-2mm}\raisebox{2mm}{\phantom{-} 0.67 }\microplot{
(1.00, 0.77)
(0.82, 0.75)
(0.83, 0.75)
(0.94, 0.77)
(0.87, 0.73)
(0.70, 0.74)
(0.79, 0.76)
(0.92, 0.76)
(0.91, 0.77)
} & \cellcolor{Brown3!24}\hspace{-2mm}\raisebox{2mm}{\phantom{-} 0.49 }\microplot{
(1.00, 0.77)
(0.82, 0.75)
(0.83, 0.75)
(0.94, 0.77)
(0.87, 0.73)
(0.70, 0.74)
(0.79, 0.76)
(0.92, 0.76)
(0.91, 0.77)
(0.96, 0.79)
(0.96, 0.79)
(0.94, 0.79)
(0.86, 0.78)
(1.00, 0.79)
(0.79, 0.79)
(0.92, 0.78)
(0.92, 0.79)
(0.95, 0.79)
(0.85, 0.79)
(1.00, 0.75)
(0.96, 0.74)
(0.99, 0.76)
(0.93, 0.75)
(0.88, 0.72)
(0.93, 0.74)
(0.89, 0.70)
(0.94, 0.72)
(0.84, 0.73)
(0.99, 0.80)
(0.96, 0.80)
(0.89, 0.80)
(0.90, 0.80)
(0.94, 0.80)
(0.87, 0.78)
(0.87, 0.75)
(0.86, 0.77)
(0.82, 0.77)
(0.84, 0.80)
}\\
\raisebox{2mm}{L2} & \cellcolor{Brown3!35}\hspace{-2mm}\raisebox{2mm}{\phantom{-} 0.71 }\microplot{
(0.97, 0.76)
(0.94, 0.76)
(0.94, 0.73)
(0.90, 0.72)
(0.92, 0.70)
(0.91, 0.69)
(0.89, 0.64)
(0.83, 0.69)
(0.97, 0.72)
(0.80, 0.66)
} & \cellcolor{Brown3!23}\hspace{-2mm}\raisebox{2mm}{\phantom{-} 0.46 }\microplot{
(0.91, 0.76)
(1.00, 0.80)
(0.81, 0.77)
(0.84, 0.79)
(0.92, 0.80)
(0.85, 0.76)
(0.86, 0.77)
(0.77, 0.76)
(0.85, 0.77)
(0.89, 0.74)
} & \cellcolor{Brown3!24}\hspace{-2mm}\raisebox{2mm}{\phantom{-} 0.47 }\microplot{
(0.88, 0.77)
(0.93, 0.78)
(0.96, 0.79)
(0.94, 0.79)
(0.92, 0.74)
(0.97, 0.79)
(0.86, 0.78)
(0.87, 0.79)
(0.81, 0.78)
(0.77, 0.75)
} & \cellcolor{Brown3!7}\hspace{-2mm}\raisebox{2mm}{\phantom{-} 0.14 }\microplot{
(0.89, 0.76)
(0.95, 0.77)
(0.81, 0.75)
(0.68, 0.75)
(0.90, 0.75)
(0.92, 0.72)
(0.88, 0.74)
(0.83, 0.74)
(0.91, 0.75)
} & \cellcolor{Brown3!22}\hspace{-2mm}\raisebox{2mm}{\phantom{-} 0.44 }\microplot{
(0.89, 0.76)
(0.95, 0.77)
(0.81, 0.75)
(0.68, 0.75)
(0.90, 0.75)
(0.92, 0.72)
(0.88, 0.74)
(0.83, 0.74)
(0.91, 0.75)
(0.88, 0.77)
(0.93, 0.78)
(0.96, 0.79)
(0.94, 0.79)
(0.92, 0.74)
(0.97, 0.79)
(0.86, 0.78)
(0.87, 0.79)
(0.81, 0.78)
(0.77, 0.75)
(0.97, 0.76)
(0.94, 0.76)
(0.94, 0.73)
(0.90, 0.72)
(0.92, 0.70)
(0.91, 0.69)
(0.89, 0.64)
(0.83, 0.69)
(0.97, 0.72)
(0.80, 0.66)
(0.91, 0.76)
(1.00, 0.80)
(0.81, 0.77)
(0.84, 0.79)
(0.92, 0.80)
(0.85, 0.76)
(0.86, 0.77)
(0.77, 0.76)
(0.85, 0.77)
(0.89, 0.74)
}\\
\raisebox{2mm}{Register} & \cellcolor{Brown3!37}\hspace{-2mm}\raisebox{2mm}{\phantom{-} 0.74 }\microplot{
(0.95, 0.69)
(0.96, 0.76)
(0.96, 0.76)
(0.98, 0.77)
} & \cellcolor{Brown3!44}\hspace{-2mm}\raisebox{2mm}{\phantom{-} 0.88 }\microplot{
(0.81, 0.80)
(0.83, 0.74)
(0.75, 0.60)
(0.77, 0.56)
(0.82, 0.80)
} & \cellcolor{Brown3!15}\hspace{-2mm}\raisebox{2mm}{\phantom{-} 0.31 }\microplot{
(0.87, 0.78)
(0.95, 0.77)
(0.95, 0.78)
(0.87, 0.77)
(0.91, 0.79)
} & \cellcolor{Blue3!8}\hspace{-2mm}\raisebox{2mm}{ -0.17 }\microplot{
(0.88, 0.70)
(0.80, 0.77)
(0.99, 0.76)
(0.88, 0.77)
(0.90, 0.75)
(0.86, 0.77)
} & \cellcolor{Brown3!22}\hspace{-2mm}\raisebox{2mm}{\phantom{-} 0.44 }\microplot{
(0.88, 0.70)
(0.80, 0.77)
(0.99, 0.76)
(0.88, 0.77)
(0.90, 0.75)
(0.86, 0.77)
(0.87, 0.78)
(0.95, 0.77)
(0.95, 0.78)
(0.87, 0.77)
(0.91, 0.79)
(0.95, 0.69)
(0.96, 0.76)
(0.96, 0.76)
(0.98, 0.77)
(0.81, 0.80)
(0.83, 0.74)
(0.75, 0.60)
(0.77, 0.56)
(0.82, 0.80)
}\\
\raisebox{2mm}{Phrasal} & \cellcolor{Brown3!25}\hspace{-2mm}\raisebox{2mm}{\phantom{-} 0.51 }\microplot{
(0.97, 0.77)
(0.95, 0.71)
(0.98, 0.77)
(0.95, 0.74)
(0.97, 0.74)
(0.95, 0.75)
(0.97, 0.73)
} & \cellcolor{Brown3!29}\hspace{-2mm}\raisebox{2mm}{\phantom{-} 0.59 }\microplot{
(0.93, 0.79)
(0.85, 0.77)
(0.85, 0.78)
(0.88, 0.80)
(0.92, 0.76)
(0.88, 0.77)
(0.81, 0.73)
} & \cellcolor{Brown3!15}\hspace{-2mm}\raisebox{2mm}{\phantom{-} 0.30 }\microplot{
(0.94, 0.79)
(0.95, 0.79)
(0.90, 0.79)
(0.99, 0.79)
(0.99, 0.80)
(0.94, 0.77)
(0.97, 0.79)
(0.85, 0.78)
} & \cellcolor{Blue3!34}\hspace{-2mm}\raisebox{2mm}{ -0.67 }\microplot{
(0.99, 0.77)
(0.87, 0.78)
(0.98, 0.75)
(0.86, 0.78)
(0.91, 0.75)
(0.89, 0.76)
(0.96, 0.75)
(0.98, 0.76)
(0.86, 0.77)
} & \cellcolor{Brown3!9}\hspace{-2mm}\raisebox{2mm}{\phantom{-} 0.18 }\microplot{
(0.99, 0.77)
(0.87, 0.78)
(0.98, 0.75)
(0.86, 0.78)
(0.91, 0.75)
(0.89, 0.76)
(0.96, 0.75)
(0.98, 0.76)
(0.86, 0.77)
(0.94, 0.79)
(0.95, 0.79)
(0.90, 0.79)
(0.99, 0.79)
(0.99, 0.80)
(0.94, 0.77)
(0.97, 0.79)
(0.85, 0.78)
(0.97, 0.77)
(0.95, 0.71)
(0.98, 0.77)
(0.95, 0.74)
(0.97, 0.74)
(0.95, 0.75)
(0.97, 0.73)
(0.93, 0.79)
(0.85, 0.77)
(0.85, 0.78)
(0.88, 0.80)
(0.92, 0.76)
(0.88, 0.77)
(0.81, 0.73)
}\\
\null\\[-0.5em]
\bf \raisebox{2mm}{All error augmenters} & \cellcolor{Brown3!38}\hspace{-2mm}\raisebox{2mm}{\phantom{-} 0.76 }\microplot{
(0.98, 0.76)
(0.88, 0.73)
(0.91, 0.73)
(0.79, 0.74)
(0.85, 0.70)
(0.86, 0.62)
(0.76, 0.61)
(0.71, 0.52)
(0.68, 0.40)
(0.70, 0.49)
(0.67, 0.63)
(0.67, 0.66)
(0.75, 0.64)
(0.67, 0.61)
(0.77, 0.70)
(0.79, 0.68)
(0.86, 0.70)
(0.77, 0.70)
(0.87, 0.69)
(0.90, 0.74)
(1.00, 0.76)
(0.95, 0.77)
(0.92, 0.74)
(0.88, 0.72)
(0.83, 0.67)
(0.87, 0.68)
(0.78, 0.59)
(0.81, 0.65)
(0.76, 0.53)
(0.76, 0.53)
(1.00, 0.75)
(0.96, 0.74)
(0.99, 0.76)
(0.93, 0.75)
(0.88, 0.72)
(0.93, 0.74)
(0.89, 0.70)
(0.94, 0.72)
(0.84, 0.73)
(0.97, 0.76)
(0.94, 0.76)
(0.94, 0.73)
(0.90, 0.72)
(0.92, 0.70)
(0.91, 0.69)
(0.89, 0.64)
(0.83, 0.69)
(0.97, 0.72)
(0.80, 0.66)
(0.95, 0.69)
(0.96, 0.76)
(0.96, 0.76)
(0.98, 0.77)
(0.97, 0.77)
(0.95, 0.71)
(0.98, 0.77)
(0.95, 0.74)
(0.97, 0.74)
(0.95, 0.75)
(0.97, 0.73)
} & \cellcolor{Brown3!35}\hspace{-2mm}\raisebox{2mm}{\phantom{-} 0.70 }\microplot{
(0.94, 0.80)
(0.85, 0.78)
(0.83, 0.78)
(0.84, 0.76)
(0.76, 0.69)
(0.68, 0.64)
(0.70, 0.65)
(0.66, 0.51)
(0.65, 0.50)
(0.80, 0.77)
(0.76, 0.77)
(0.78, 0.77)
(0.62, 0.74)
(0.74, 0.77)
(0.72, 0.76)
(0.72, 0.78)
(0.60, 0.75)
(0.75, 0.77)
(0.76, 0.75)
(0.99, 0.80)
(0.96, 0.80)
(0.71, 0.71)
(0.78, 0.77)
(0.80, 0.77)
(0.72, 0.75)
(0.71, 0.77)
(0.79, 0.72)
(0.57, 0.47)
(0.74, 0.69)
(0.99, 0.80)
(0.96, 0.80)
(0.89, 0.80)
(0.90, 0.80)
(0.94, 0.80)
(0.87, 0.78)
(0.87, 0.75)
(0.86, 0.77)
(0.82, 0.77)
(0.84, 0.80)
(0.91, 0.76)
(1.00, 0.80)
(0.81, 0.77)
(0.84, 0.79)
(0.92, 0.80)
(0.85, 0.76)
(0.86, 0.77)
(0.77, 0.76)
(0.85, 0.77)
(0.89, 0.74)
(0.81, 0.80)
(0.83, 0.74)
(0.75, 0.60)
(0.77, 0.56)
(0.82, 0.80)
(0.93, 0.79)
(0.85, 0.77)
(0.85, 0.78)
(0.88, 0.80)
(0.92, 0.76)
(0.88, 0.77)
(0.81, 0.73)
} & \cellcolor{Brown3!21}\hspace{-2mm}\raisebox{2mm}{\phantom{-} 0.42 }\microplot{
(0.99, 0.79)
(0.87, 0.76)
(0.90, 0.77)
(0.93, 0.78)
(0.81, 0.77)
(0.76, 0.69)
(0.78, 0.73)
(0.85, 0.78)
(0.78, 0.64)
(0.79, 0.59)
(0.94, 0.78)
(0.79, 0.73)
(0.80, 0.72)
(0.74, 0.70)
(0.84, 0.78)
(0.88, 0.77)
(0.80, 0.75)
(0.92, 0.78)
(0.86, 0.76)
(0.82, 0.74)
(0.76, 0.75)
(0.95, 0.79)
(0.84, 0.75)
(0.95, 0.79)
(0.90, 0.79)
(0.84, 0.74)
(0.84, 0.78)
(0.83, 0.71)
(0.83, 0.74)
(0.84, 0.60)
(0.96, 0.79)
(0.96, 0.79)
(0.94, 0.79)
(0.86, 0.78)
(1.00, 0.79)
(0.79, 0.79)
(0.92, 0.78)
(0.92, 0.79)
(0.95, 0.79)
(0.85, 0.79)
(0.88, 0.77)
(0.93, 0.78)
(0.96, 0.79)
(0.94, 0.79)
(0.92, 0.74)
(0.97, 0.79)
(0.86, 0.78)
(0.87, 0.79)
(0.81, 0.78)
(0.77, 0.75)
(0.87, 0.78)
(0.95, 0.77)
(0.95, 0.78)
(0.87, 0.77)
(0.91, 0.79)
(0.94, 0.79)
(0.95, 0.79)
(0.90, 0.79)
(0.99, 0.79)
(0.99, 0.80)
(0.94, 0.77)
(0.97, 0.79)
(0.85, 0.78)
} & \cellcolor{Brown3!12}\hspace{-2mm}\raisebox{2mm}{\phantom{-} 0.24 }\microplot{
(0.94, 0.75)
(0.85, 0.74)
(0.94, 0.76)
(0.75, 0.76)
(0.72, 0.68)
(0.92, 0.74)
(0.68, 0.71)
(0.72, 0.69)
(0.78, 0.63)
(0.70, 0.64)
(0.90, 0.74)
(0.81, 0.73)
(0.76, 0.73)
(0.80, 0.75)
(0.85, 0.75)
(0.88, 0.75)
(0.87, 0.75)
(0.80, 0.74)
(0.78, 0.71)
(0.87, 0.76)
(0.82, 0.76)
(0.94, 0.76)
(0.78, 0.75)
(0.94, 0.77)
(0.88, 0.75)
(0.83, 0.74)
(0.81, 0.66)
(0.77, 0.64)
(0.84, 0.61)
(0.83, 0.65)
(1.00, 0.77)
(0.82, 0.75)
(0.83, 0.75)
(0.94, 0.77)
(0.87, 0.73)
(0.70, 0.74)
(0.79, 0.76)
(0.92, 0.76)
(0.91, 0.77)
(0.89, 0.76)
(0.95, 0.77)
(0.81, 0.75)
(0.68, 0.75)
(0.90, 0.75)
(0.92, 0.72)
(0.88, 0.74)
(0.83, 0.74)
(0.91, 0.75)
(0.88, 0.70)
(0.80, 0.77)
(0.99, 0.76)
(0.88, 0.77)
(0.90, 0.75)
(0.86, 0.77)
(0.99, 0.77)
(0.87, 0.78)
(0.98, 0.75)
(0.86, 0.78)
(0.91, 0.75)
(0.89, 0.76)
(0.96, 0.75)
(0.98, 0.76)
(0.86, 0.77)
} & \cellcolor{Brown3!27}\hspace{-2mm}\raisebox{2mm}{\phantom{-} 0.53 }\microplot{
(0.98, 0.76)
(0.88, 0.73)
(0.91, 0.73)
(0.79, 0.74)
(0.85, 0.70)
(0.86, 0.62)
(0.76, 0.61)
(0.71, 0.52)
(0.68, 0.40)
(0.70, 0.49)
(0.94, 0.80)
(0.85, 0.78)
(0.83, 0.78)
(0.84, 0.76)
(0.76, 0.69)
(0.68, 0.64)
(0.70, 0.65)
(0.66, 0.51)
(0.65, 0.50)
(0.99, 0.79)
(0.87, 0.76)
(0.90, 0.77)
(0.93, 0.78)
(0.81, 0.77)
(0.76, 0.69)
(0.78, 0.73)
(0.85, 0.78)
(0.78, 0.64)
(0.79, 0.59)
(0.94, 0.75)
(0.85, 0.74)
(0.94, 0.76)
(0.75, 0.76)
(0.72, 0.68)
(0.92, 0.74)
(0.68, 0.71)
(0.72, 0.69)
(0.78, 0.63)
(0.70, 0.64)
(0.67, 0.63)
(0.67, 0.66)
(0.75, 0.64)
(0.67, 0.61)
(0.77, 0.70)
(0.79, 0.68)
(0.86, 0.70)
(0.77, 0.70)
(0.87, 0.69)
(0.90, 0.74)
(0.80, 0.77)
(0.76, 0.77)
(0.78, 0.77)
(0.62, 0.74)
(0.74, 0.77)
(0.72, 0.76)
(0.72, 0.78)
(0.60, 0.75)
(0.75, 0.77)
(0.76, 0.75)
(0.94, 0.78)
(0.79, 0.73)
(0.80, 0.72)
(0.74, 0.70)
(0.84, 0.78)
(0.88, 0.77)
(0.80, 0.75)
(0.92, 0.78)
(0.86, 0.76)
(0.82, 0.74)
(0.90, 0.74)
(0.81, 0.73)
(0.76, 0.73)
(0.80, 0.75)
(0.85, 0.75)
(0.88, 0.75)
(0.87, 0.75)
(0.80, 0.74)
(0.78, 0.71)
(0.87, 0.76)
(1.00, 0.76)
(0.95, 0.77)
(0.92, 0.74)
(0.88, 0.72)
(0.83, 0.67)
(0.87, 0.68)
(0.78, 0.59)
(0.81, 0.65)
(0.76, 0.53)
(0.76, 0.53)
(0.99, 0.80)
(0.96, 0.80)
(0.71, 0.71)
(0.78, 0.77)
(0.80, 0.77)
(0.72, 0.75)
(0.71, 0.77)
(0.79, 0.72)
(0.57, 0.47)
(0.74, 0.69)
(0.76, 0.75)
(0.95, 0.79)
(0.84, 0.75)
(0.95, 0.79)
(0.90, 0.79)
(0.84, 0.74)
(0.84, 0.78)
(0.83, 0.71)
(0.83, 0.74)
(0.84, 0.60)
(0.82, 0.76)
(0.94, 0.76)
(0.78, 0.75)
(0.94, 0.77)
(0.88, 0.75)
(0.83, 0.74)
(0.81, 0.66)
(0.77, 0.64)
(0.84, 0.61)
(0.83, 0.65)
(1.00, 0.75)
(0.96, 0.74)
(0.99, 0.76)
(0.93, 0.75)
(0.88, 0.72)
(0.93, 0.74)
(0.89, 0.70)
(0.94, 0.72)
(0.84, 0.73)
(0.99, 0.80)
(0.96, 0.80)
(0.89, 0.80)
(0.90, 0.80)
(0.94, 0.80)
(0.87, 0.78)
(0.87, 0.75)
(0.86, 0.77)
(0.82, 0.77)
(0.84, 0.80)
(0.96, 0.79)
(0.96, 0.79)
(0.94, 0.79)
(0.86, 0.78)
(1.00, 0.79)
(0.79, 0.79)
(0.92, 0.78)
(0.92, 0.79)
(0.95, 0.79)
(0.85, 0.79)
(1.00, 0.77)
(0.82, 0.75)
(0.83, 0.75)
(0.94, 0.77)
(0.87, 0.73)
(0.70, 0.74)
(0.79, 0.76)
(0.92, 0.76)
(0.91, 0.77)
(0.97, 0.76)
(0.94, 0.76)
(0.94, 0.73)
(0.90, 0.72)
(0.92, 0.70)
(0.91, 0.69)
(0.89, 0.64)
(0.83, 0.69)
(0.97, 0.72)
(0.80, 0.66)
(0.91, 0.76)
(1.00, 0.80)
(0.81, 0.77)
(0.84, 0.79)
(0.92, 0.80)
(0.85, 0.76)
(0.86, 0.77)
(0.77, 0.76)
(0.85, 0.77)
(0.89, 0.74)
(0.88, 0.77)
(0.93, 0.78)
(0.96, 0.79)
(0.94, 0.79)
(0.92, 0.74)
(0.97, 0.79)
(0.86, 0.78)
(0.87, 0.79)
(0.81, 0.78)
(0.77, 0.75)
(0.89, 0.76)
(0.95, 0.77)
(0.81, 0.75)
(0.68, 0.75)
(0.90, 0.75)
(0.92, 0.72)
(0.88, 0.74)
(0.83, 0.74)
(0.91, 0.75)
(0.95, 0.69)
(0.96, 0.76)
(0.96, 0.76)
(0.98, 0.77)
(0.81, 0.80)
(0.83, 0.74)
(0.75, 0.60)
(0.77, 0.56)
(0.82, 0.80)
(0.87, 0.78)
(0.95, 0.77)
(0.95, 0.78)
(0.87, 0.77)
(0.91, 0.79)
(0.88, 0.70)
(0.80, 0.77)
(0.99, 0.76)
(0.88, 0.77)
(0.90, 0.75)
(0.86, 0.77)
(0.97, 0.77)
(0.95, 0.71)
(0.98, 0.77)
(0.95, 0.74)
(0.97, 0.74)
(0.95, 0.75)
(0.97, 0.73)
(0.93, 0.79)
(0.85, 0.77)
(0.85, 0.78)
(0.88, 0.80)
(0.92, 0.76)
(0.88, 0.77)
(0.81, 0.73)
(0.94, 0.79)
(0.95, 0.79)
(0.90, 0.79)
(0.99, 0.79)
(0.99, 0.80)
(0.94, 0.77)
(0.97, 0.79)
(0.85, 0.78)
(0.99, 0.77)
(0.87, 0.78)
(0.98, 0.75)
(0.86, 0.78)
(0.91, 0.75)
(0.89, 0.76)
(0.96, 0.75)
(0.98, 0.76)
(0.86, 0.77)
}\\
\bottomrule\end{tabular}
 \\[-0.5em]
\caption{
Pearson correlation within each error type and prompt (averaged across models and languages).
See Appendix \Cref{fig:beryllium} for visualization in a single plot.
}
\label{tab:boron}
\end{table*}

Further, the best-performing prompt for one model can be the worst-performing prompt for another, which is in line with prior findings \citep{voronov-etal-2024-mind}.
We observed that GPT-4o tends to benefit from the ``\texttt{\#\#\#}'' structure of Prompt 1 while the other models tend to copy parts of this prompt regardless of error type or intensity, leading to lower scores.
Similarly, prompt 4 works best of all prompts for EuroLLM and Qwen 2.5, but makes Gemini produce more off-target translations.

The results suggest that \textbf{users can benefit from choosing the `right' prompt} for the model they are using, with shorter prompts being somewhat preferable. %
This seems to be \textbf{at least as important as avoiding certain error types} (see also \Cref{subsec:compare-noise}).

\subsection{Comparing Error Types}
\label{subsec:compare-noise}

On average across all prompts, the baseline uniform error augmenter affects the translation quality the most.
This suggests that models may be naturally less robust to unrealistic errors due to their exposure to realistic errors during training.

The simple spelling transformations (common errors and phonetically inspired misspelling) capture one of the errors commonly made by L2 speakers.
The spelling transformations have a strong effect on the LLMs' performance, both in terms of similarity measures to the original prompt as well as a low translation quality. 

Phrase-level simplifications show the least overall effect: For Prompt 4, we even see a negative correlation, meaning some substitutions performed better than the original prompt.
Thus, while lack of fluency on the part of L2 users may introduce unnaturalness or awkwardness as perceived by native speakers, the models may actually respond well to this simplification.
The same effect also holds for the register simplifications.

The `L2' and `Lazy User' scenarios model natural compositions of error types that L2 users or laypeople respectively might display. Specifically, these scenarios combine the orthographic errors with phrasal and register errors, respectively.
The `L2' scenario shows similar rates of impact as the orthographic errors alone, possibly due to the simplifying effect of phrasal errors as discussed above.
The `Lazy User' scenario, however, shows more impact, i.e. rate of degradation per unit error, than its constituent error types, indicating that the presence of the different error types has a compounding effect on the impact of each on model performance.

\subsection{Frequency of Off-Target Translations} 

\begin{figure*}[htbp]
    \centering
    \includegraphics[width=1\linewidth]{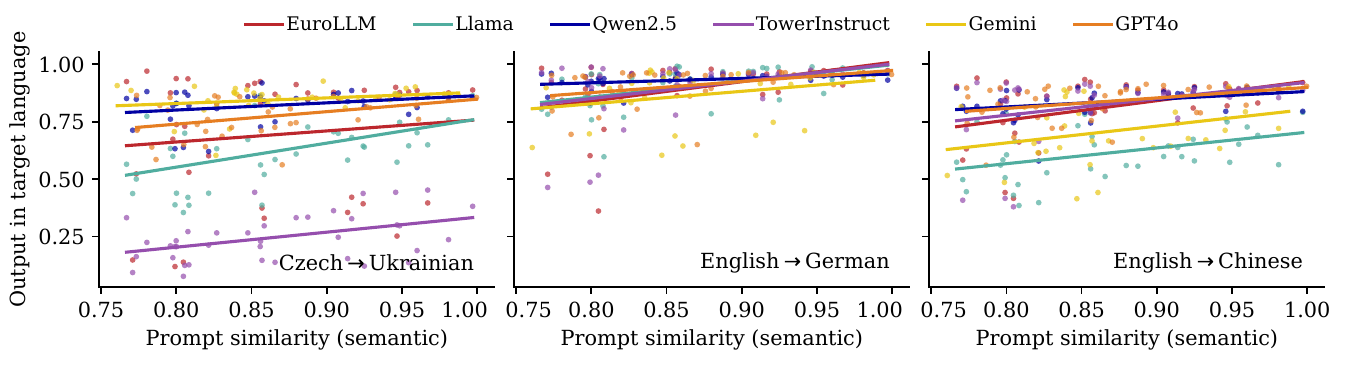}
    \caption{Percentage of outputs in the target language, by language pair and model.
    Note that TowerInstruct does not officially support Ukrainian or Czech.}

    \label{fig:on-target-percent}
\end{figure*}

A common failure mode for LLMs-as-MTs is responding in a non-target language. 
\Cref{fig:on-target-percent} shows the proportion of outputs in the target language by model and language pair.
\textbf{For all models and language pairs, high error intensities decrease the proportion of on-target outputs.}
Of the target languages, German has the highest proportion of on-target outputs.
For Czech-Ukrainian, TowerInstruct tends to output other languages even with the original prompt, because the two languages are not well-supported by the model.

Note also that COMET does not penalize wrong language output and may still score off-target outputs highly, for instance, if the model outputs Russian translation when Ukrainian translation is requested.

\begin{figure}[t]
\centering
\includegraphics[width=1\linewidth]{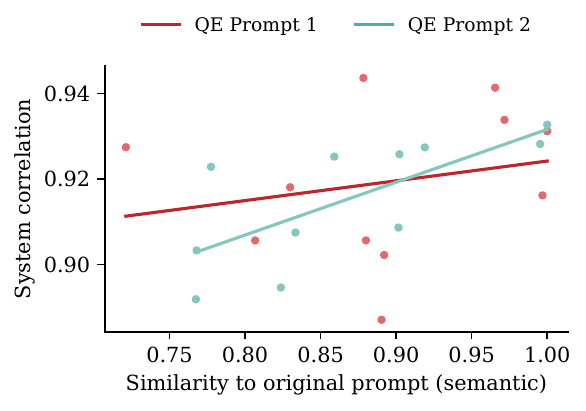}
\caption{Changing model performance, in terms of system-level correlation (y-axis), across quality estimation prompts augmented with orthographic errors, against semantic similarity of the error-augmented prompt to its original (x-axis). The results suggest only a weak trend.}
\label{fig:quality-v-sim-qe}
\end{figure}

\subsection{Transferability to Quality Estimation}

\Cref{fig:quality-v-sim-qe} shows that both quality estimation prompts are (weakly) affected by applying \textit{realistic} orthographic errors, with differences in both the base prompts' performance and the effect of the error.
This reinforces that prompt choice matters for quality estimation as for translation.

\newcommand{\microplotqesys}[1]{
\begin{tikzpicture}
  \begin{axis}[
    xmin=0.6, xmax=1.05,
    ymin=0.75, ymax=1.0,
    ticks=none,
    axis x line=bottom,
    axis y line=left,
    enlargelimits=false,
    height=2.5cm,
    width=2.5cm,
  ]
    \addplot[only marks, mark=*, mark size=0.8pt, color=black] coordinates {
      #1
    };
  \end{axis}
\end{tikzpicture}
}

\newcommand{\microplotqeseg}[1]{
\begin{tikzpicture}
  \begin{axis}[
    xmin=0.6, xmax=1.05,
    ymin=0.0, ymax=0.7,
    ticks=none,
    axis x line=bottom,
    axis y line=left,
    enlargelimits=false,
    height=2.5cm,
    width=2.5cm,
  ]
    \addplot[only marks, mark=*, mark size=0.8pt, color=black] coordinates {
      #1
    };
  \end{axis}
\end{tikzpicture}
}
\begin{table}[t]
\small \centering
\begin{tabular}{ll@{\hspace{2mm}}l@{\hspace{2mm}}l}\toprule
\bf Level & \bf Prompt 1 & \bf Prompt 2 & \bf All prompts\\
\midrule
\raisebox{2mm}{System-} & \cellcolor{Brown3!8}\hspace{-2mm}\raisebox{2mm}{\phantom{-} 0.15 }\microplotqesys{
(0.81, 0.91)
(0.88, 0.91)
(0.88, 0.94)
(0.97, 0.94)
(0.83, 0.92)
(1.00, 0.92)
(0.72, 0.93)
(0.97, 0.93)
(0.89, 0.89)
(0.89, 0.90)
} & \cellcolor{Brown3!36}\hspace{-2mm}\raisebox{2mm}{\phantom{-} 0.71 }\microplotqesys{
(0.77, 0.89)
(1.00, 0.93)
(1.00, 0.93)
(0.82, 0.89)
(0.90, 0.91)
(0.77, 0.90)
(0.78, 0.92)
(0.90, 0.93)
(0.83, 0.91)
(0.86, 0.93)
(0.92, 0.93)
} & \cellcolor{Brown3!22}\hspace{-2mm}\raisebox{2mm}{\phantom{-} 0.43 }\microplotqesys{
(0.81, 0.91)
(0.88, 0.91)
(0.88, 0.94)
(0.97, 0.94)
(0.83, 0.92)
(1.00, 0.92)
(0.72, 0.93)
(0.97, 0.93)
(0.89, 0.89)
(0.89, 0.90)
(0.77, 0.89)
(1.00, 0.93)
(1.00, 0.93)
(0.82, 0.89)
(0.90, 0.91)
(0.77, 0.90)
(0.78, 0.92)
(0.90, 0.93)
(0.83, 0.91)
(0.86, 0.93)
(0.92, 0.93)
}\\
\raisebox{2mm}{Segment-} & \cellcolor{Blue3!19}\hspace{-2mm}\raisebox{2mm}{ -0.38 }\microplotqeseg{
(0.81, 0.16)
(0.88, 0.17)
(0.88, 0.16)
(0.97, 0.16)
(0.83, 0.16)
(1.00, 0.16)
(0.72, 0.17)
(0.97, 0.15)
(0.89, 0.15)
(0.89, 0.16)
} & \cellcolor{Brown3!28}\hspace{-2mm}\raisebox{2mm}{\phantom{-} 0.57 }\microplotqeseg{
(0.77, 0.14)
(1.00, 0.17)
(1.00, 0.17)
(0.82, 0.16)
(0.90, 0.16)
(0.77, 0.15)
(0.78, 0.18)
(0.90, 0.17)
(0.83, 0.15)
(0.86, 0.16)
(0.92, 0.17)
} & \cellcolor{Brown3!5}\hspace{-2mm}\raisebox{2mm}{\phantom{-} 0.09 }\microplotqeseg{
(0.81, 0.16)
(0.88, 0.17)
(0.88, 0.16)
(0.97, 0.16)
(0.83, 0.16)
(1.00, 0.16)
(0.72, 0.17)
(0.97, 0.15)
(0.89, 0.15)
(0.89, 0.16)
(0.77, 0.14)
(1.00, 0.17)
(1.00, 0.17)
(0.82, 0.16)
(0.90, 0.16)
(0.77, 0.15)
(0.78, 0.18)
(0.90, 0.17)
(0.83, 0.15)
(0.86, 0.16)
(0.92, 0.17)
}\\
\null\\[-1em]
\bottomrule\end{tabular}
 \\[-0.5em]
\caption{
Pearson correlation between system and segment-level evaluation performance and prompt similarity, for the orthographic error and quality estimation prompts (averaged across languages).
}
\label{tab:boron-qe}
\end{table}

\Cref{tab:boron-qe} shows Pearson correlations of system-level correlations with the prompt similarity, per prompt.
A higher correlation implies that decreasing prompt similarity also decreases quality estimation correlation with human judgments.
We observe a similar effect of orthographic errors on the system-level correlation of GEMBA across languages, with an overall correlation of 0.43, compared to 0.49 for translation.
This suggests the effect of orthographic errors is transferable and largely consistent across tasks.
See \ACref{fig:qe-lithium-lang-results} for per-language results.

However, we observe a negative correlation for Quality Estimation Prompt 1 at the segment level.
The original, error-free prompt already achieves a poor segment-level correlation of 0.16.
This may be an artifact of our strict setting which prohibits retries and artificially sets the resulting score to 0 to elucidate the error impact. 
Further, the shorter prompt may explain the reduced variance in outputs and therefore weaker correlations, though additional testing is required to elucidate this effect.

\subsection{Qualitative Analysis}
We performed qualitative analysis on the machine translation outputs by manually inspecting a sample of the lowest-scoring translations in each setting.
We also sampled ten source segments per language pair with their translations from every setting, to understand how various error types and levels change the translation of a given sentence.

\paragraph{Models add supplementary information.}
In addition to providing the translation, Gemini frequently offers useful background information, such as multiple versions of the translation or the pronunciation of the Chinese translation explained in Latin script.
This insight explains why Gemini produces longer outputs than any other model, as shown in \Cref{fig:lengths}.
The extra information appears in the outputs regardless of the error type or intensity.
On the other hand, the model explains its choice of words more frequently with increasing error intensity.
This behavior may be beneficial for a user, but is difficult to parse in an automated setting.

\paragraph{Lower scores are often due to worse instruction following.}
The translations by a single model tend to remain relatively stable across various error types and levels.
The main explanation for the differing scores is the presence of redundant text alongside the translation, such as adding the name of the target language, saying ``here is your translation,'' repeating the source sentence, or paraphrasing the prompt. 
The base prompts in our experiments either explicitly state to only output the translation and nothing else, or strongly imply it by their structure.
Therefore, when models output redundant text rather than only providing the translation as instructed, we consider it worse instruction following.

GPT-4o is a notable exception: It generates more diverse translations and less redundant text, unless subjected to a high level of uniform errors that render the text unreadable to humans ($p > 0.5$).
Up to that threshold, a lower metric score for GPT-4o is more likely to correspond to genuinely lower translation quality.

This finding is also supported quantitatively in \Cref{fig:lengths}.
It shows that adding errors increases the average length of the LLM output and that there are frequently significant differences between target and reference length.
GPT-4o produces the shortest texts while Gemini produces the longest.
The average length of Qwen2.5 outputs is the least affected by errors. 

\paragraph{LLMs can still translate with illegible prompts.}
Realistic noising scenarios produce prompts that are mostly legible to humans.
We also stress-tested the models by applying uniform errors with $p > 0.46$, exceeding the natural error range.
This transformation makes the prompts largely illegible to humans (compare \Cref{tab:noised_prompts}), for example: ``\textit{Reaajaky fgo trormm \{src\_lang\} ttk \{tgt\_lang\}:: \textbackslash n \{src\_lang\} \{src\_text\} \textbackslash n \{tgt\_lang\}::}''

LLMs sometimes produce an error message or request clarification without providing a translation in response to an illegible prompt.
However, they frequently produce valid translations even when given prompts with a high $p\geq0.7$, which are nonsensical to the human eye.
These translations are frequently accompanied by strategies such as copying the prompt verbatim, attempting to translate or fix it, or treating the text as a cipher to decode.
In rare cases, they ignore the errors altogether and only provide the expected translation.
We show examples of these outputs in Appendix \Cref{tab:llm_output_examples}.

The only parts of the prompt that remain legible in this stress-testing scenario are the unchanged source and target languages, as well as the source sentence to be translated.
This suggests that if an LLM is capable of performing a task, it recognizes the task based on subtle hints and perform it even with an objectively bad prompt.
This finding may be helpful for future research, as it implies that if an LLM does not generalize to a task as demonstrated by a handful of prompts, further prompt engineering efforts are unlikely to change that outcome.

\vspace{-2mm}
\section{Conclusion}
\vspace{-2mm}

We investigated the impact of imperfect prompt construction on LLM performance.
We model a range of user-inspired errors and apply them to prompts in a controlled manner, observing the resulting degradation in task performance over 6 models, 2 tasks, and 4 language pairs. 
We find that spelling errors have the most severe effect on performance, while phrase-level disfluencies and simplifications have lower impact and may even help in some cases.
We also explore natural compositions of error types, finding that these compound the effects of their constituent error types.
A qualitative analysis of the resulting outputs reveals that `imperfect' English in prompts often does not lead to lower translation quality, but rather worse instruction following.
This usually demonstrates as the model not performing the task, or performing additional tasks on top of translation, such as attempting to fix the errors in the text, providing multiple variants of the translation, or explaining the translation word by word.
This makes the output more difficult to parse automatically and reflects negatively in the automatic evaluation; however, a user would be able to extract the translation in the cases where it is provided.

Crucially, the effect of the initial prompt selection is greater than that of the majority of realistic user errors, emphasizing the importance of prompt choice. 
Yet, while practitioners may therefore benefit from optimizing prompt selection over a set of diverse and error-free prompts, lay users are unlikely to do so, and may further exhibit errors in their prompts due to imperfect language skills or other reasons.
This work highlights the gap between LLM performance as evaluated in pristine conditions as opposed to real-world conditions, given a diverse user base, and calls for improving LLM resilience to prompt choice as well as user errors in prompts.

\section*{Limitations}

This study only looks at errors in English prompts for MT-related tasks; error-classes as well as the nature of impact of typical user errors in different language prompts may naturally differ.

We used automatically generated errors rather than using error data from real learners.
One important reason for this is the difficulty of sourcing real examples.
While using generated errors may mean that some of the examples are less realistic, it allows for a broader statistical analysis and provides us with better control over our experimental variables.

\section*{Ethics Statement}

We do not anticipate any negative ethical implications arising from this study.
We took care to ensure realistic representations of errors without casting users in a negative light.

The total inference cost for the two proprietary models (GPT-4o-mini and Gemini-2.0-flash) is less than USD 100.
While we did run the other models locally, the overall cost for all the models likely does not exceed USD 200.

The licenses for the open-weights models are: Llama 3.1 Community License for Llama 3.1; Apache 2.0 for EuroLLM and the Qwen model we used; and CC-BY-NC-4.0 for the Tower model we used (with its base model Llama 2 being licensed under the Llama 2 Community License).
These licenses all permit our use of the model weights.

We used AI-assisted coding (i.e. Copilot) with the bulk being human-written. For writing, AI was used to check grammar mistakes.

\bibliography{misc/anthology.min.bib,misc/bibliography.bib}
\bibliographystyle{misc/acl_natbib}

\clearpage

\appendix
\section{Descriptions of Noising Functions}\label{app:noise-functions}

\subsection{Uniform errors}
\label{app:subsec:random-noise}

This error augmenter, parameterized by probability $p$, introduces random perturbations into the prompt, modeling natural typos. 
The perturbations include random character transposition, omission, doubling, and substitution for neighbouring letters on the keyboard.
The character-wise frequency of error is controlled by $p$, and the type of error is sampled uniformly from the error types.

\subsection{Orthographic error}
\label{app:subsec:ortho-noise}

This error augmenter models spelling errors, both due to imperfect proficiency in written English as well as random typos.
\citet{cook1997l2} provide a classification of the types of spelling errors made by both L1 and L2 speakers, and report the relative frequency of these errors, finding higher error rates for L2 speakers, but similar distributions over error categories.
Guided by this work, we define the follow classes of orthographic errors:
\begin{itemize}[left=0mm,noitemsep,topsep=0mm]
    \item \textbf{Natural typos}: We re-use our uniform error augmenter as described above. This corresponds to the category ``other'' as defined by \citet{cook1997l2}.
    \item \textbf{Omission}: Omitting one of a non-word-initial consonant pair (e.g. \texttt{ck$\rightarrow$k}), dropping \texttt{r} before a consonant, dropping \texttt{e} if it is word-final, or before \texttt{ly}.
    \item \textbf{Insertion}: Doubling non-word-initial consonant.
    \item \textbf{Substitution}: Confusing specific sets of consonants (such as \texttt{s}, \texttt{c}, \texttt{z}), confusing vowels with each other. For the latter, we generate errors consistently with the finding that confusions between \texttt{a}, \texttt{e}, \texttt{i} constitute 60\% of vowel substitutions.
    \item \textbf{Transposition}: Transposing consecutive vowels (\texttt{ie$\rightarrow$ei}), transposing certain bigrams (\texttt{er}, \texttt{ng}).
\end{itemize}

\noindent
Similarly to the uniform errors, the orthographic error augmenter is controlled by a parameter $p$, which corresponds to the probability of error on a given character. 
Varying $p$ allows us therefore to generate prompts over different error intensities.
Given a character to be perturbed, we sample a type of error from the above list, as per the natural distribution over these categories of error described in \citet{cook1997l2}.
Given a type of error (e.g. substitution), we uniformly sample a subtype of error from all subtypes applicable to the character and its context.
For example, the ``a-e''-confusion subtype is only relevant for ``a's''. 
Note that a character may have no relevant subtypes under a given type: In this case, we simply skip the character.

\subsection{Phonetic LLM-generated errors}
\label{app:subsec:phonetic-llm}

We also investigate the impact on LLM performance of errors made by non-native speakers writing English sentences based on phonetic transcriptions in their first languages. We prompt an LLM to mimic these errors in various languages (Arabic, Chinese, German, Polish and Spanish) spoken by beginner English learners. According to our tests, LLMs can simulate typical phonetic errors for a particular language, despite not being fully fluent in it. Example for a Polish person: \emph{Translate the following line from English to Chinese.} $\rightarrow$ \emph{Translejt de follouing lajn from English tu Chinese.}

\subsection{Phrasal simplification}
\label{app:subsec:phrasal-noise}

We would like to study the effect of alternate lexical/phrasal simplification, as possibly committed by L2 speakers.
Note that prompts generally use largely restricted vocabulary, and potential phrasal errors are therefore limited.
We consider two levels of L2 proficiency: Beginner and intermediate, and prompt an LLM to mimic such errors made by L2 speakers of each level, generating $k=10$ error-augmented candidates per prompt and level.
We manually examine the generations and discard implausible options.
We find that LLM-generated errors cover a reasonable range of plausible errors of this type.

\subsection{Register changes}\label{app:subsec:register-noise}

We are also interested in the effect of informal registers of users, who may query LLMs similarly to querying search engines, with non-standard casing, dropping of articles and function words, and reframing for conciseness.
For example, \textit{Translate from de to en}$\rightarrow$\textit{translate de - en}.
This type of errors also offers a limited number of possible transformations of a base prompt.
Similarly to above, we prompt an LLM to generate $k=10$ informal versions of each base prompt with the above changes, for two levels (medium and high) of informality, and manually discard unlikely candidates.

\section{Implementation Details}\label{app:eval-details}
\label{sec:implementation_details}

\vspace{-2mm}
For evaluation we use the following settings:
\begin{itemize}[topsep=0mm]
\item ChrF: {\small nrefs:1|case:mixed|eff:yes|nc:6|nw:0|space:no|version:2.3.1} \citep[sacrebleu]{post-2018-call}
\item COMET: {\small Python3.11.5|Comet2.2.5|fp32|Unbabel/wmt22-comet-da|r1} (\citealp{rei-etal-2022-comet}, sacrecomet \citealp{zouhar-etal-2024-pitfalls})
\end{itemize}

\section{Ablations}

\subsection{Impact of User Errors Per Language}
\label{sec:impact_language}

\begin{figure*}[htbp]
    \centering
    \includegraphics[width=0.95\linewidth]{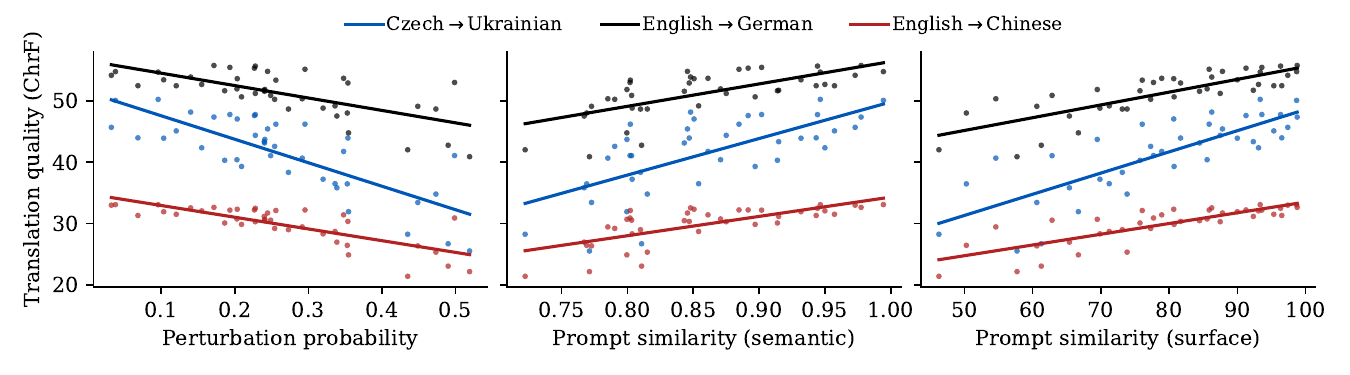}
    \includegraphics[width=0.95\linewidth]{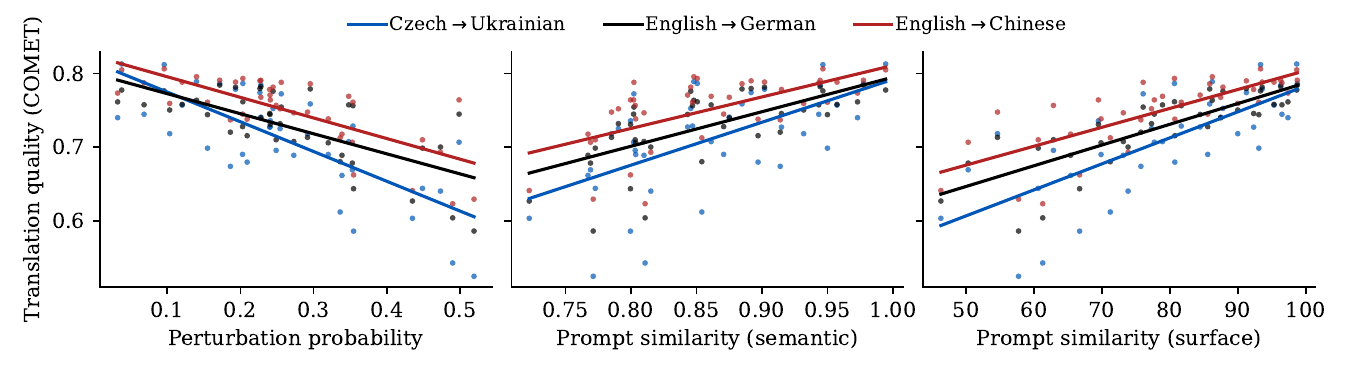}
    \caption{Sensitivity to error augmentation by language pair.
    Translation quality measured by ChrF (top) or COMET (bottom), given a certain amount of error augmentation (x-axes). 
    The error augmentation probability refers to the probability $p$ of applying orthographic errors. Prompt similarity (semantic) refers to the inner product of sentence embeddings. Prompt similarity (surface) refers to the chrF score of the error-augmented prompt against the base prompt.}
    \label{fig:per-language-sensitivity}
\end{figure*}

We further examine how different language pairs and models respond to user errors.
\ACref{fig:per-language-sensitivity} shows orthographic error sensitivity per language pair averaged across all models.
We show six subplots, one for each combination of quality metric (ChrF, COMET) and measure of error level (noising probability $p$, semantic prompt similarity, and surface prompt similarity). 

\begin{figure*}[htbp]
    \centering
    \includegraphics[width=1\linewidth]{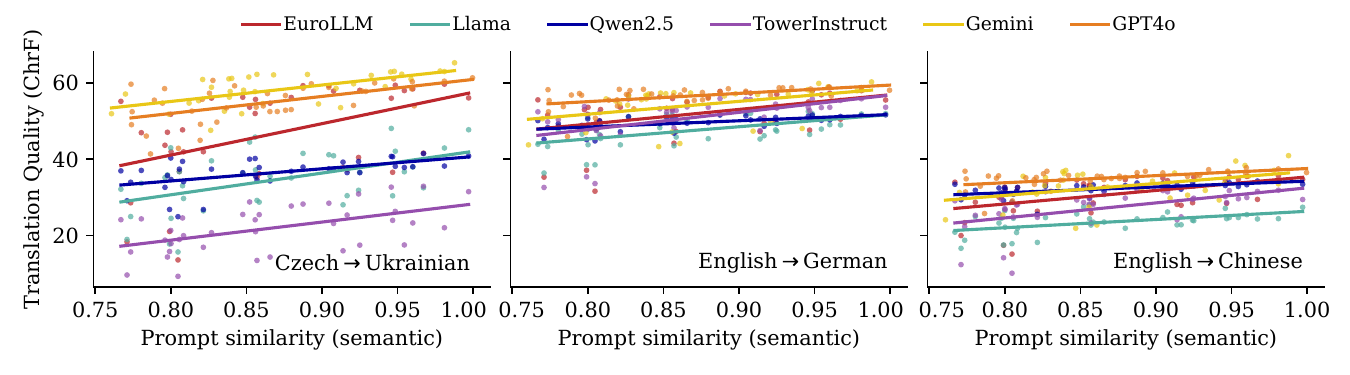}
    \includegraphics[width=1\linewidth]{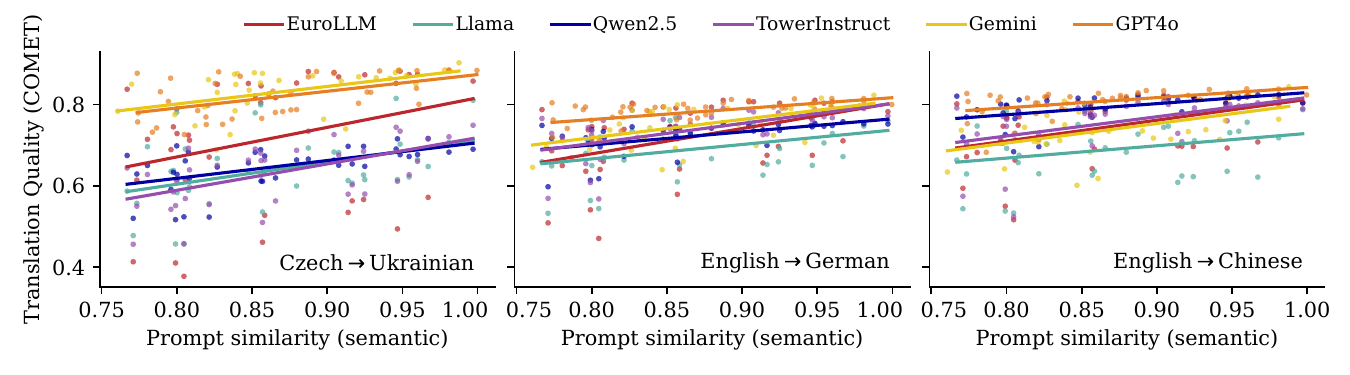}
    \caption{Sensitivity of individual models to prompt noising, for each language pair and by model.
    x-axis: Prompt similarity to base prompt (semantic). y-axes: Translation quality measured by ChrF (top) and COMET (bottom).}
    \label{fig:quality-by-model}
  
\end{figure*}

In general, all language pairs are affected to a similar degree.
Czech-Ukrainian appears slightly more sensitive than the other two, possibly due to less robust support of the models for this language pair, while translation into Chinese scores lower on ChrF.
Similarly, \ACref{fig:quality-by-model} shows the sensitivity per language pair \emph{and model}.
All models are affected, for all languages, to a similar degree.
Note that TowerInstruct does not support Czech or Ukrainian, and Llama-3.1 officially does not support Czech, Ukrainian, or Chinese.

\begin{table*}[htbp]
    \small \centering
    \begin{tabular}{p{16cm}}
    \toprule
    \textbf{Prompt 1:} \#\#\# Instruction:\textbackslash n Translate Input from \{src\_lang\} to \{tgt\_lang\} \textbackslash n \#\#\# Input:\textbackslash n \{src\_text\}\textbackslash n \#\#\# Response:\textbackslash n \\
    \textbf{Prompt 2:} Translate the following line from\textbackslash n \{src\_lang\} to \{tgt\_lang\}.\textbackslash n Be very literal, and only translate the content of the line, do not add any explanations: \{src\_text\} \\
     \textbf{Prompt 3:} Translate this from \{src\_lang\} to \{tgt\_lang\}:\textbackslash n \{src\_lang\}: \{src\_text\}\textbackslash n \{tgt\_lang\}: \\
     \textbf{Prompt 4:} Translate the following text from \{src\_lang\} to \{tgt\_lang\}.\textbackslash n \{src\_text\}  \\
     
    \textbf{Prompt minimal:} \{src\_lang\}: \{src\_text\}\textbackslash n \{tgt\_lang\}: \\
     \bottomrule
    \end{tabular}
    \caption{Base forms for investigated machine translation prompts.}
    \label{tab:base-prompts}
\end{table*}

\begin{table*}[htbp]
    \small \centering
    \begin{tabular}{p{16cm}}
    \toprule
    \textbf{QE Prompt 1:} Score the following translation from \{src\_lang\} to \{tgt\_lang\} on a continuous scale from 0 to 100, where a score of zero means `no meaning preserved' and score of one hundred means `perfect meaning and grammar'.\textbackslash n \{src\_lang\} source: `\{src\_text\}'\textbackslash n \{tgt\_lang\} translation: `\{tgt\_text\}'\textbackslash n Score:  \\
    \textbf{QE Prompt 2:} Please analyze the given source and translated sentences and output a translation quality score on a continuous scale ranging from 0 to 100. Translation quality should be evaluated based on both fluency and adequacy. A score close to 0 indicates a low quality translation, while a score close to 100 indicates a high quality translation. Do not provide any explanations or text apart from the score.\textbackslash n \{src\_lang\} Sentence: \{src\_text\}\textbackslash n \{tgt\_lang\} Sentence: \{tgt\_text\}\textbackslash n Score: \\
     \bottomrule
    \end{tabular}
    \caption{Base forms for investigating quality estimation prompts.}
    \label{tab:qe-base-prompts}
\end{table*}

\begin{figure*}
    \centering
    \includegraphics[width=1\linewidth]{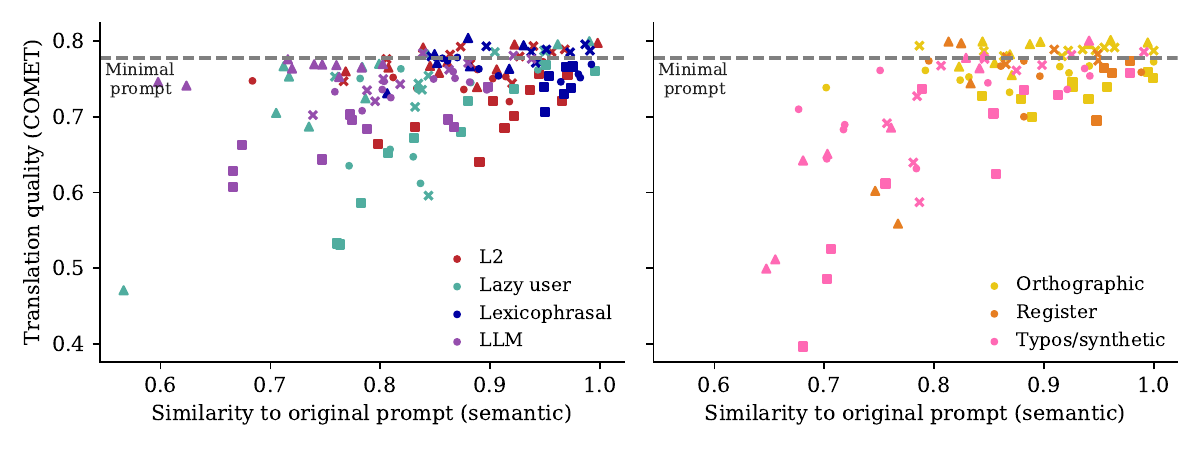}
    \caption{
    Average performance (across models and languages) with respect to individual prompts and error types.
    Each shape is one of four prompts.
    Visualizes \Cref{tab:boron} in a single plot.
    }
    \label{fig:beryllium}
\end{figure*}

\begin{figure*}[htbp]
    \centering
    \includegraphics[width=\linewidth]{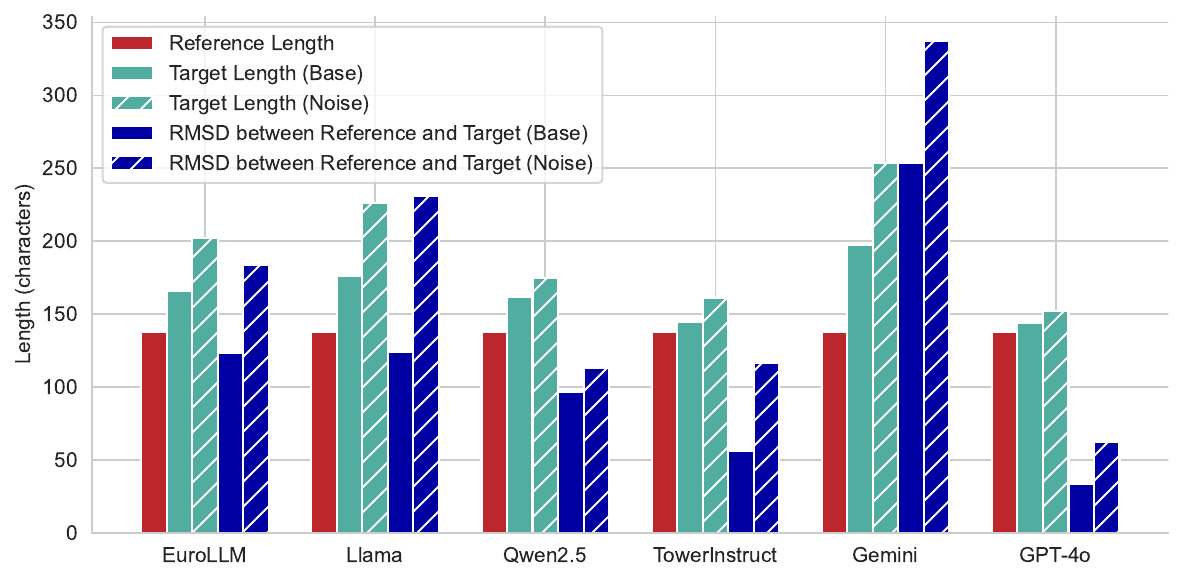}
    \caption{Average length of generated outputs compared to the reference length. We compare the output lengths given the base prompts to the aggregate of all error augmenter variants, and observe that error-augmented prompts lead to consistently longer texts.}
    \label{fig:lengths}
\end{figure*}

\begin{table*}[t]
    \small  \centering
    \begin{tabular}{@{\hspace{1mm}}p{15.3cm}@{\hspace{1mm}}}
    \toprule
        \textbf{EuroLLM, Orthographic (0.7):}
        Rrwnslswfee tn gplowwgn linnee ffro \textbackslash n English t German.\textbackslash n Bee yg oiiteerarl, ann olyn rtnlqata thet cnotetn kff feg linns, ri ohtt adad anay exlqntaiiox:: \textit{Ich finde es sehr lebensbejahend; aus so einfachen Zutaten etwas so Reichhaltiges und Komplexes zu machen. Wie bei der Geburt eines Kindes.}

        \textbf{Qwen2.5, Orthographic (0.8):}
        Hier ist die Übersetzung von Englisch ins Deutsche:  \textit{Offensichtlich gibt es nur zwei Ebenen, die im Original-Super-Mario-Maker noch nicht abgeschlossen wurden.}
        \\
        
        \textbf{Gemini, Orthographic (0.8):}
        Gerne, hier ist die Übersetzung des Satzes "I better get some motorbikes fixed, or else!" ins Deutsche:\textbackslash n \textbackslash n Ich sollte besser ein paar Motorräder reparieren lassen, oder aber!
        \\
        
        \textbf{Llama, Orthographic (0.8):}
        The encoded message is a cipher. It appears to be a simple substitution cipher, where each letter has been shifted by a certain number of positions in the alphabet. After analyzing the text, I believe it is a Caesar cipher with a shift of 13 positions.  Here's the decoded message:  Original: Gafnsnlatae tet ffllwlif llnie gon Decoded:  "Fantastical life line gone" [...] \\

        \textbf{GPT-4o, Orthographic (0.9):}
        It seems like there are some typos and formatting issues in your message. However, I can help you translate the provided English text into German. Here’s the translation:  **English:** "I find it very life affirming; transforming such simple ingredients into something so rich and complex. Like witnessing a birth."  **German:** \textit{"Ich finde es sehr lebensbejahend; so einfache Zutaten in etwas so Reichhaltiges und Komplexes zu verwandeln. Wie das Zeugnis einer Geburt."}  If you need further assistance or corrections, feel free to ask! \\
        \textbf{TowerLLM, Orthographic (0.8):} Gafnsnlatae tet ffllwlif llnie gon \textbackslash n Deutsch gehen Englisch.\textbackslash n Sehr traurig, wenn man bedenkt, dass die Mieten in den letzten Jahren um durchschnittlich 6 Prozent pro Jahr gestiegen sind, aber diese Vorschläge würden die Mieten um bis zu 15 Prozent erhöhen, was ironischerweise höher ist als die historischen jährlichen Preissteigerungen. [...]. \\
    \bottomrule      
    \end{tabular}
    \caption{Examples of LLM outputs when presented with perturbed prompts.}
    \label{tab:llm_output_examples}
\end{table*}

\begin{figure*}[htbp]
    \centering
    \includegraphics[width=1.0\linewidth]{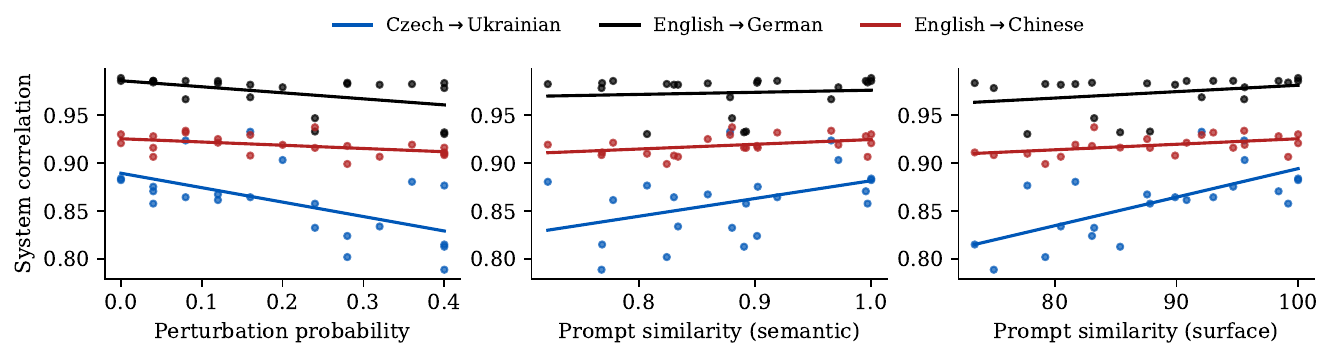}
    \caption{Sensitivity of QE outputs to perturbations by language pair. System-level correlation is measured against human evaluations on the test set, given a certain perturbation amount. Semantic prompt similarity measures the inner product of erroneous and base sentence embeddings, while surface similarity measures ChrF between erroneous and base prompts. The results show that effects are seen across language pairs, though the magnitude of the effect varies.}
    \label{fig:qe-lithium-lang-results}
\end{figure*}

\newcommand{\checkyes}{\includegraphics[height=1.2em]{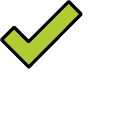}}
\newcommand{\checknoo}{\includegraphics[height=1.2em]{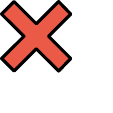}}
\newcommand{\checkque}{\includegraphics[height=1.2em]{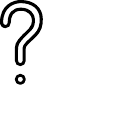}}

\begin{table*}[ht]
    \centering
    \small
    \begin{tabular}{l lllll |@{\hspace{7mm}}l}\toprule
        & \bf English & \bf Czech & \bf Ukrainian & \bf German & \bf Chinese & \bf Open\\
        \midrule
        Claude-3.5 Sonnet$^1$ & \checkyes & \checkque & \checkque & \checkque & \checkque & \checknoo \\
        CommandR+ & \checkyes & \checkyes$^2$ & \checkyes$^2$ & \checkyes & \checkyes & \checknoo \\
        GPT-4o & \checkyes & \checkyes & \checkyes & \checkyes & \checkyes & \checknoo \\
        Gemini-1.5 Pro & \checkyes & \checkyes & \checkyes & \checkyes & \checkyes & \checknoo \\
        Phi-3 & \checkyes & \checkyes & \checkyes & \checkyes & \checkyes & \checkyes \\
        Phi-4 14B & \checkyes & \checkyes & \checkyes & \checkyes & \checkyes & \checkyes \\
        EuroLLM & \checkyes & \checkyes & \checkyes & \checkyes & \checkyes & \checkyes \\
        Llama & \checkyes & \checknoo & \checknoo & \checkyes & \checknoo & \checkyes \\
        Tower & \checkyes & \checknoo$^3$ & \checknoo$^3$ & \checkyes & \checkyes & \checkyes \\
        Aya23 & \checkyes & \checkyes & \checkyes & \checkyes & \checkyes & \checkyes \\
        DeepSeek-V3$^1$ & \checkyes & \checkque & \checkque & \checkque & \checkyes & \checkyes \\
        Qwen-2.5 & \checkyes & \checknoo & \checknoo & \checknoo & \checknoo & \checkyes \\
        Mistral & \checkyes & \checknoo & \checknoo & \checkyes & \checkyes & \checkyes \\        
    \bottomrule
    \end{tabular}
    \caption{List of models taken into consideration. The list of supported languages for the open-weight models is taken from their Hugging Face model cards. \\
        $^1$: The model is multilingual but the list of supported languages is not available;\\
        $^2$: Languages included in the pre-training but not post-training (\href{https://docs.cohere.com/v2/docs/command-r-plus\#multilingual-capabilities}{Cohere documentation}); \\    
        $^3$: Tower70B took part to WMT2024 on the Czech$\rightarrow$Ukrainian language pair \cite{kocmi-etal-2024-findings}, but the model card for \href{https://huggingface.co/Unbabel/TowerInstruct-7B-v0.2}{Unbabel/TowerInstruct-7B-v0.2} does not include it.
    }
    \label{tab:model-selection}
\end{table*}

\end{document}